\documentclass[11pt]{article}

\usepackage[preprint]{acl}
\usepackage{url}
\usepackage{xurl}

\usepackage{adjustbox} % for max width box
\usepackage{colortbl}
\usepackage{xcolor}
\usepackage{placeins}
\usepackage{hyperref}
\usepackage{amsmath}
\usepackage{booktabs}
\usepackage{tabularx}
\usepackage{array}
\usepackage{makecell}
\usepackage{amssymb}
\definecolor{hlgreen}{HTML}{D5F5D5}   % light green for improvements
\definecolor{hlred}{HTML}{FDDEDE}     % light red for degradation
\newcolumntype{C}{>{\centering\arraybackslash}X}
\newcommand{\NA}{\mbox{--}}
\newcommand{\tight}{\setlength{\tabcolsep}{3.5pt}} % tighten tables a bit
\usepackage{times}
\usepackage{latexsym}
\usepackage[T1]{fontenc}
\usepackage[utf8]{inputenc}

\usepackage{microtype}

\usepackage{inconsolata}
\usepackage{multirow}

\usepackage{graphicx}
\newcommand\blfootnote[1]{%
  \begingroup\renewcommand\thefootnote{}\footnote{#1}%
  \addtocounter{footnote}{-1}\endgroup}
\title{MedRAGChecker: Claim-Level Verification for Biomedical Retrieval-Augmented Generation}

\author{
  Yuelyu Ji \quad Min Gu Kwak \quad Hang Zhang \quad Xizhi Wu \quad Chenyu Li \quad Yanshan Wang \\
  University of Pittsburgh, Pittsburgh, PA, USA \\
  \texttt{yueluji@gmail.com}
}

\begin{document}
\maketitle
\blfootnote{Code, prompts, and diagnostics available at: 
\url{https://github.com/JoyDajunSpaceCraft/MedicalRagChecker}}
\begin{abstract}
Biomedical retrieval-augmented generation (RAG) produces long-form 
answers that mix supported, under-evidenced, and contradicted claims,
demanding fine-grained verification. We propose \textsc{MedRAGChecker},
a claim-level diagnostic framework that (i) decomposes answers into 
atomic claims, (ii) verifies each claim by combining textual natural 
language inference (NLI) with a knowledge-graph (KG) consistency 
signal, and (iii) uses a \emph{selective gating} rule to fuse the two 
signals only when the KG signal is decisive, falling back to NLI 
otherwise. To assess our framework on biomedical RAG, we distill a 
GPT-4.1 teacher into compact student checkers and evaluate across 
four biomedical QA benchmarks (PubMedQA, MedQuAD, LiveQA, MedRedQA) 
with three biomedical generators. \textsc{MedRAGChecker} achieves 
comparable answer-level faithfulness to general-purpose RAG evaluators 
while additionally producing safety-critical error diagnostics that 
align with human judgment on decision-flip claims (69.8\% agreement 
vs.\ 63.4\% for NLI-only).
As an additional analysis, we instantiate the KG component with two 
back-ends of differing scope---DRKG and BioPortal---and find that 
KG coverage substantially affects fusion quality: an incomplete KG 
can inflate hallucination via false contradictions on uncovered 
claims, motivating our selective gating design.
\end{abstract}
\begin{figure}[t]
\centering
\includegraphics[width=\columnwidth]{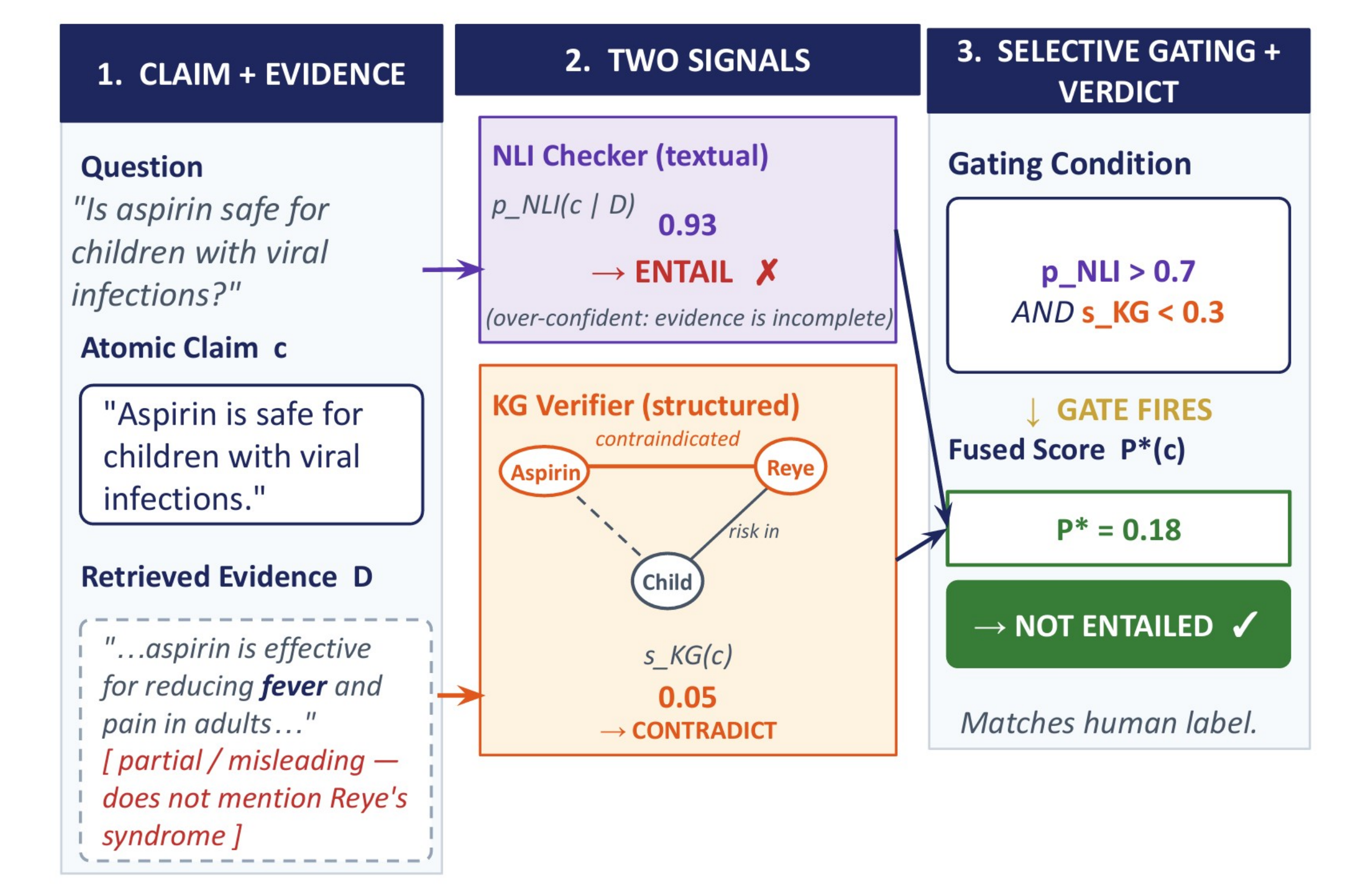}
\caption{\textbf{Selective gating in \textsc{MedRAGChecker}.}
For the atomic claim ``Aspirin is safe for children with
viral infections,'' the textual NLI checker is over-confident
in entailment ($p_{\mathrm{NLI}}{=}0.93$) because the retrieved
evidence does not mention Reye's syndrome. The KG verifier
returns low support ($s_{\mathrm{KG}}{=}0.05$). The selective
gating condition (high $p_{\mathrm{NLI}}$, low $s_{\mathrm{KG}}$)
fires, fusing the two signals into $P^\star{=}0.18$ and
producing the correct \textsc{Not Entailed} verdict that matches
the human label.}
\label{fig:intro}
\end{figure}

% However,

% \begin{figure*}[t]
% e~\ref{fig:overall}). Our method first breaks a candidate answer into atomic claims (e.g., splitting ``Chronic aspirin use reduces stroke risk'' into concrete subclaims) and then verifies each claim using two complementary signals.

% , and we show that a simple F1-weighted ensemble over students yields more reliable \textsc{Entail}/\textsc{Neutral}/\textsc{Contradict} predictions than any single model.

\section{Introduction}

Large language models (LLMs) now pass standardized medical exams,
yet their ungrounded generations still produce clinically harmful
errors~\cite{yu2025large,singhal2025toward,vladika2024healthfc,pandit2025medhallu}.
Retrieval-augmented generation (RAG) partly addresses this by
conditioning answers on up-to-date medical
literature~\cite{achiam2023gpt,tu2024towards,lewis2020retrieval},
but noisy retrieval can still cause the model to anchor on
misleading snippets~\cite{ning2025medtrust}. Long-form biomedical
answers contain many distinct factual statements, and whole-answer
scoring hides isolated but clinically important mistakes. This
motivates \emph{claim-level} evaluation: decomposing each answer
into atomic factual claims (e.g., a single
drug--effect or drug--contraindication assertion) and verifying
each one independently~\cite{min2023factscore,ru2024ragchecker}.

Standard claim-level verification frames each claim as a natural
language inference (NLI) hypothesis against retrieved evidence.
In biomedicine, however, textual entailment alone is insufficient:
a claim can be linguistically supported by retrieved passages yet
violate known drug--disease contraindications. Recent work therefore
augments NLI with biomedical knowledge graphs
(KGs)~\cite{vladika2024healthfc,ning2025medtrust}, under the
implicit assumption that adding a KG can only help. \emph{This
assumption is unsafe.} Figure~\ref{fig:intro} illustrates the
failure: a claim about ibuprofen's gastrointestinal side effects
receives a spurious \textsc{Contradict} verdict from a KG that
lacks the relevant entity pair, yet is correctly entailed by a
broader-coverage KG. Whether a KG helps depends on whether it
covers the claims being verified --- and no prior work has
measured how this coverage drives verification quality on
biomedical RAG.

\begin{figure*}[t]
\centering
\includegraphics[width=\linewidth]{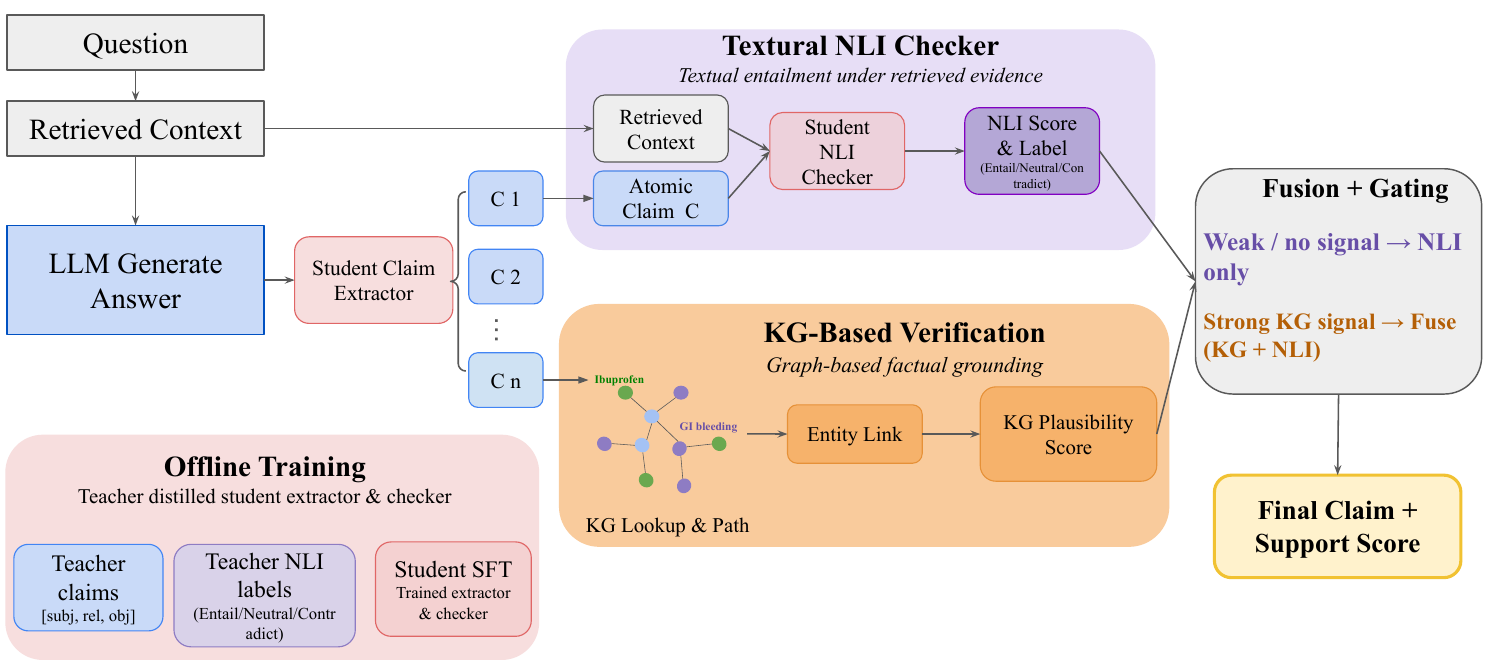}
\caption{\textbf{Overview of \textsc{MedRAGChecker}.} For each atomic
claim extracted from the answer, a textual NLI checker (purple) and a
KG-based verifier (orange) produce complementary signals. These are
combined by a selective gating module: when the KG provides a strong
signal of conflict, the KG score is fused with the NLI score; otherwise
the NLI score is used alone. The student NLI checker and claim extractor
are distilled from a GPT-4.1 teacher offline (dashed arrows); at
inference no teacher calls are made. KG coverage, shown later to be
the dominant factor in verification quality (Section~\ref{sec:rq_coverage}),
is varied across two back-ends: DRKG and BioPortal.}
\label{fig:overall}

\end{figure*}

We propose \textsc{MedRAGChecker} (Figure~\ref{fig:overall}), a
claim-level diagnostic framework that combines NLI verification
with a KG plausibility signal under a coverage-aware fusion rule.
To isolate the effect of coverage from confounding algorithmic
choices, we instantiate the framework with two KGs that differ
primarily in scope: a drug-repurposing KG
(DRKG~\cite{drkg2020}, 62.7\% entity coverage on our benchmarks)
and a multi-ontology hub
(BioPortal~\cite{salvadores2013bioportal}, 94.2\% coverage). The
31.5pp coverage gap drives a 9.0pp swing in hallucination rate
(34.2\% vs.\ 25.2\%). To turn this finding into a practical fix,
we introduce \emph{selective gating}: applying the KG signal only
when it strongly conflicts with a confident NLI entailment, the
narrow regime where a false KG signal would do the most damage.
Beyond what individual components contribute, we additionally 
observe that the choice of KG back-end---specifically its 
coverage over the target claim distribution---substantially 
affects fusion quality. This motivates a coverage-aware fusion 
rule (selective gating) that remains robust across KG choices.

% \paragraph{Contributions.}
% \textbf{(i)}~We identify a failure mode of KG-augmented biomedical
% RAG verification: incomplete KGs \emph{inflate} hallucination rate
% through false contradictions on uncovered claims, rather than
% reducing it. \textbf{(ii)}~We introduce selective gating, a
% coverage-aware fusion rule that fires on ${\sim}12.3\%$ of
% KG-linked claims and concentrates correction on safety-critical
% drug--disease and drug--adverse-event relations. \textbf{(iii)}~We
% validate across four biomedical QA benchmarks, three generators,
% and two KGs, with a targeted human study on decision-flip claims
% showing that coverage-aware fusion raises agreement with
% annotators from 63.4\% to 69.8\%.
\paragraph{Contributions.}
\textbf{(i)}~We propose \textsc{MedRAGChecker}, a claim-level 
biomedical RAG diagnostic framework that combines distilled NLI 
checkers with a teacher-independent KG verifier under a selective 
gating rule, and demonstrate that it matches general-purpose 
RAG evaluators on faithfulness while uniquely surfacing 
safety-critical errors.
\textbf{(ii)}~We introduce \emph{selective gating}, a fusion 
rule that applies KG adjustment only when NLI is confident of 
entailment and KG is confident of conflict, firing on 
${\sim}12.3\%$ of KG-linked claims and concentrating correction 
on safety-critical relations.
\textbf{(iii)}~Through an analysis with two KG back-ends 
(DRKG, BioPortal), we show that KG coverage substantially 
affects fusion quality, validated across four biomedical QA 
benchmarks with a targeted human study where coverage-aware 
fusion raises decision-flip agreement from 63.4\% to 69.8\%.

% TODO contratict

\section{Related Work}
% TODO 
% \paragraph{Biomedical QA.}
% Biomedical QA benchmarks (e.g., MedQA, MedMCQA, PubMedQA) drove domain-specific LMs, and systems such as Med-PaLM2 achieved strong performance on standardized medical exams \cite{jin2020disease,pmlr-v174-pal22a,jin-etal-2019-pubmedqa}.
% However, long-form biomedical answers remained prone to hallucinations \cite{asgari2025framework,huang2025proactive}, making safety-critical validation necessary.
\paragraph{Biomedical QA.}
Despite strong performance on standardized exams~\cite{jin2020disease,pmlr-v174-pal22a,jin-etal-2019-pubmedqa}, long-form biomedical answers remain prone to hallucinations~\cite{asgari2025framework,huang2025proactive}, making claim-level validation necessary.
% MedRAGChecker addressed this by verifying fine-grained claims rather than scoring answers holistically, targeting better diagnostic capabilities for biomedical responses.
\paragraph{RAG Evaluation and Fact-Checking.}
Prior work formulates RAG factuality checking as an NLI-style claim-
evidence problem, splitting generations into statements and verifying
each against retrieved
passages~\cite{lewis2020retrieval,fu2024gptscore,xin2025improving}.
RAGAS~\citep{es-etal-2024-ragas} and
RAGChecker~\citep{ru2024ragchecker} evaluate retrieval and grounding
at the \emph{answer level}. \textsc{MedRAGChecker} operates at the
\emph{claim level} and aggregates verified claims into answer-level
diagnostics, adding a structured KG signal for claim veracity.
HealthFC~\cite{vladika2024healthfc} offers a strong clinical baseline
but targets short, well-formed statements, while we target multi-
sentence RAG answers.
\paragraph{Knowledge Graphs for QA.}
Biomedical KGs support QA reasoning~\cite{himmelstein2017systematic}, but their incompleteness limits standalone use~\cite{chen2022explainable}. We use KGs as external verifiers complementing text-based NLI, fusing both signals to improve robustness on safety-critical claims.
\section{MedRAGChecker}
\label{sec:framework}

% Figure~\ref{fig:overall} overviews the pipeline. MedRAGChecker combines a textual NLI signal from teacher-distilled student checkers with a KG-based consistency signal from DRKG, and fuses them into $P^\star(c_i)$ used by all downstream diagnostics.
Figure~\ref{fig:overall} overviews the pipeline. \textsc{MedRAGChecker} combines a textual NLI signal from teacher-distilled student checkers with a KG-based consistency signal (instantiated with DRKG and BioPortal), and fuses them into a single calibrated support score used by all downstream diagnostics.

\subsection{Task Definition and Outputs}
\label{subsec:task}
 
We cast biomedical RAG evaluation as claim-centric verification.
Given a RAG instance $(q, D, a)$ with question $q$, generated answer
$a$, and top-$k$ retrieved passages $D=\{d_j\}_{j=1}^{k}$,
\textsc{MedRAGChecker} decomposes $a$ into atomic claims
$\mathcal{C}=\{c_1,\dots,c_n\}$ and assigns each claim
(1)~a discrete verdict
$\hat y_i \in \{\textsc{Entail},\textsc{Neutral},\textsc{Contradict}\}$
and (2)~a calibrated support score $s(c_i) \in [0,1]$.
Two signals produce the verdict: a textual NLI probability
$p_{\mathrm{NLI}}(c_i)$ against retrieved evidence, and a KG
plausibility score $s_{\mathrm{KG}}(c_i)$ against a biomedical
knowledge base. We study how the choice of KG, and specifically its
\emph{coverage} over claims in our benchmarks, affects the quality of
the final verdict.
 
In our implementation, the textual checker provides
$p_{\mathrm{NLI}}(c_i) \triangleq P(\textsc{Entail}\mid c_i, D)$ from
the ensemble distribution. When KG evidence is available, we fuse
this textual signal with KG support to obtain the final support score
(Section~\ref{sec:method_fusion}).

\subsection{Claim Extraction}
\label{sec:method_extract}
We use GPT-4.1 as a teacher to decompose each answer $a$ into atomic
claims, represented as subject--predicate--object (SPO) triples
(e.g., \texttt{["Ibuprofen", "may cause", "GI bleeding"]}), in the
spirit of FActScore-style atomic decomposition~\citep{min2023factscore}.
The triple form aligns with downstream KG entity linking
(\S\ref{sec:method_kg}).
% We use GPT-4.1 as a teacher to decompose each answer $a$ into atomic
% claims following FActScore-style evaluation~\citep{min2023factscore}.
To avoid teacher calls at inference time, we distill a biomedical LLM
(e.g., Med42-Llama3-8B) via supervised fine-tuning on
$(q, a) \rightarrow c_1, \ldots, c_n$ with teacher claim lists as targets.
Since claim extraction is a standard and solved step in this framework,
we treat it as front-end infrastructure rather than a contribution.
Extraction quality (BERTScore vs.\ teacher claims, human Likert
ratings) is validated in Section~\ref{sec:sanity_check}; training
details are in Appendix~\ref{app:student_configs};
qualitative error analysis is in Appendix~\ref{app:extraction_errors}.

\subsection{Textual NLI Verification}
\label{sec:method_nli}
 
For each extracted claim $c_i$, a checker predicts
$y_i \in \{\textsc{Entail}, \textsc{Neutral}, \textsc{Contradict}\}$
with evidence $D$ as premise and the claim as hypothesis, producing
$p_{\mathrm{NLI}}(c_i) \triangleq P(\textsc{Entail} \mid c_i, D)$.
We distill GPT-4.1 verdicts into Meditron3-8B, Med-Qwen2-7B, and
Med42-Llama3-8B via supervised fine-tuning (SFT), then combine them in an F1-weighted ensemble
so minority-class specialists contribute more to \textsc{Contradict}
decisions. Unless otherwise stated, all downstream diagnostics use
this ensemble; details are in Appendix~\ref{app:student_configs}.

% \subsection{KG-Based Verification}
% \label{sec:method_kg}
 
% In biomedical settings, a claim may be linguistically entailed by
% the retrieved text yet clinically wrong. For example, a claim that
% ``aspirin is safe for children with viral infections'' can be
% supported by a poorly-retrieved passage but violates the
% Aspirin--Reye contraindication. We therefore augment the textual
% NLI signal with structured biomedical knowledge. As argued in the
% introduction, the utility of a KG depends critically on whether it
% covers the claims being verified, and we design our KG
% verification stage to make this coverage effect explicit and
% measurable.

% This KG verification stage is also teacher-independent: it operates
% on extracted SPO triples without GPT-4.1 calls, providing an
% external anchor that partially decouples our framework from
% teacher bias (cf.\ Limitations).
\subsection{KG-Based Verification}
\label{sec:method_kg}

In biomedical settings, a claim may be linguistically entailed by 
the retrieved text yet clinically wrong. For example, a claim that 
``aspirin is safe for children with viral infections'' can be 
supported by a poorly-retrieved passage but violates the 
Aspirin--Reye contraindication. We therefore augment the textual 
NLI signal with structured biomedical knowledge.

Our KG verification module is designed as a generic interface with 
three stages: (1)~\emph{entity linking} from the extracted SPO triple 
to KG entities; (2)~\emph{plausibility scoring} over linked entities; 
and (3)~\emph{signal fusion} with the textual NLI score 
(Section~\ref{sec:method_fusion}). The module is teacher-independent: 
it operates on extracted SPO triples without GPT-4.1 calls, providing 
an external anchor that partially decouples our framework from 
teacher bias (cf.\ Limitations). Different KG back-ends can be 
plugged into this interface; we instantiate the module with two 
biomedical KGs in our experiments (Section~\ref{sec:exp_kg_backends}).

\subsubsection{Entity Linking}
\label{sec:method_kg_link}
Unlike general-purpose KGs (e.g., Wikidata~\cite{vrandevcic2014wikidata}, ConceptNet~\cite{liu2004conceptnet}) that ship 
with high-quality entity embeddings, the biomedical KGs we use 
either do not provide pretrained entity embeddings aligned with 
our claim vocabulary (BioPortal) or provide only relation-specific 
TransE embeddings without natural-language surface forms (DRKG). 
We therefore link entities via text-based matching enhanced with 
biomedical entity representations (SapBERT) where applicable.

For each claim $c$, we first extract an SPO triple $(s, r, o)$ from the
teacher decomposition. We then link $s$ and $o$ to the target KG.
 
For \textbf{DRKG}, we perform string-based matching against DRKG entity
names and map canonical claim relations (e.g., \textsc{treats},
\textsc{causes}) to DRKG edge types, computing a normalized string
similarity score $s_{\text{text}}(c) \in [0,1]$.
 
For \textbf{BioPortal}, we query the BioPortal API across entity-type-
appropriate ontologies, scoring each candidate entity by
\begin{equation}
\label{eq:bioportal_link}
s_{\text{link}} = 0.3 \cdot s_{\text{text}} + 0.5 \cdot s_{\text{SapBERT}} + 0.2 \cdot s_{\text{syn}},
\end{equation}
where $s_{\text{text}}$ is Jaccard string similarity,
$s_{\text{SapBERT}}$ is the cosine similarity between
SapBERT~\citep{liu2021sapbert} biomedical entity embeddings, and
$s_{\text{syn}}$ is synonym-set match. We retain the top-scoring entity per mention with
$s_{\text{link}} \ge 0.5$; otherwise the claim is treated as
KG-uncovered and falls back to NLI-only.

\subsubsection{KG Plausibility Scoring}
\label{sec:method_kg_score}
 
Given linked entities, we compute a claim-level plausibility score
$s_{\mathrm{KG}}(c) \in [0, 1]$.
 
\noindent\textbf{DRKG (TransE).} For each aligned triple $(h, r, t)$,
we compute
$p_{\mathrm{KGE}}(h, r, t) = \sigma(-\|\mathbf{e}_h + \mathbf{r}_r - \mathbf{e}_t\|_2)$
using pre-trained translation-based KG embeddings (TransE) ~\citep{bordes2013translating},
taking the maximum over candidate triples. We combine this with
text alignment:
\begin{equation}
s_{\mathrm{KG}}^{\text{DRKG}}(c)
= (1{-}\alpha) \cdot p_{\mathrm{KGE}}(c) + \alpha \cdot s_{\text{text}}(c),
\label{eq:kg_score_drkg}
\end{equation}
with $\alpha$ tuned on dev.
 
\noindent\textbf{BioPortal (ontology-based).} For each linked entity
pair, we compute the maximum of three signals: (i) \emph{direct
property lookup} (score 0.9, e.g., \texttt{indication},
\texttt{contraindication}); (ii) \emph{hierarchical proximity} via
shared ontology ancestors (0.2--0.7); and (iii) \emph{cross-ontology
mapping} through LOOM links (0.8):
\begin{equation}
s_{\mathrm{KG}}^{\text{Bio}}(c) = \max(s_{\text{direct}}, s_{\text{hier}}, s_{\text{loom}}).
\label{eq:kg_score_bio}
\end{equation}
Scoring details are in Appendix~\ref{app:bioportal_details}.
 \subsection{Signal Fusion with Selective Gating}
\label{sec:method_fusion}
 
\subsubsection{Logit-Space Fusion}
 
For claims linked to the KG, we combine NLI and KG signals in logit
space:
\begin{equation}
\label{eq:pstar_fusion}
P^\star(c_i) = \sigma\!\big(
\beta \cdot \mathrm{logit}(p_{\mathrm{NLI}}(c_i))
+ (1{-}\beta) \cdot \mathrm{logit}(s_{\mathrm{KG}}(c_i))
\big),
\end{equation}
where $\beta$ controls the relative weight. Logit-space fusion avoids
the scale mismatch between the two signals and is insensitive to
monotonic rescalings.
For claims that cannot be linked to the KG, we fall back to
$P^\star(c_i) = p_{\mathrm{NLI}}(c_i)$.
 
%  \subsubsection{Why Selective Gating Is Needed}
 
% Vanilla fusion (Eq.~\ref{eq:pstar_fusion}) treats every KG-linked
% claim symmetrically, so the KG signal always contributes weight
% $1{-}\beta$. This fails when KG coverage is imperfect or when the KG
% signal is weak but non-zero (e.g., a spurious hierarchical match for
% a lifestyle claim unrelated to any drug or disease). A low
% $s_{\mathrm{KG}}$ then forces the claim toward \textsc{Contradict}
% even when NLI strongly entails. DRKG's 57\% coverage produces a long
% tail of weak KG signals that systematically inflate hallucination,
% matching the failure mode in Figure~\ref{fig:intro}.

\subsubsection{Selective Gating}
\label{sec:method_gating}
A symmetric application of Eq.~\ref{eq:pstar_fusion} fails when 
$s_{\mathrm{KG}}$ is weak but non-zero (e.g., a spurious hierarchical 
match on a lifestyle claim), pushing the claim toward \textsc{Contradict} 
even when NLI strongly entails. We therefore apply KG adjustment only 
when both signals are decisive; with shorthand 
$\ell_{\mathrm{NLI}} \triangleq \mathrm{logit}(p_{\mathrm{NLI}})$, 
$\ell_{\mathrm{KG}} \triangleq \mathrm{logit}(s_{\mathrm{KG}})$:
% A direct application of Eq.~\ref{eq:pstar_fusion} to every KG-linked 
% claim treats the KG signal symmetrically across all claims. This 
% becomes problematic when the KG signal is weak but non-zero (e.g., 
% a spurious hierarchical match for a lifestyle claim unrelated to 
% any drug or disease): a low $s_{\mathrm{KG}}$ then pushes the claim 
% toward \textsc{Contradict} even when NLI strongly entails. We 
% therefore apply KG-based adjustment only when the KG signal indicates 
% high-confidence conflict with a strong NLI entailment; otherwise we 
% retain the NLI-only score:
% We introduce shorthand 
% $\ell_{\mathrm{NLI}}(c_i) \triangleq \mathrm{logit}(p_{\mathrm{NLI}}(c_i))$ 
% and 
% $\ell_{\mathrm{KG}}(c_i) \triangleq \mathrm{logit}(s_{\mathrm{KG}}(c_i))$ 
% to keep the expression compact:

\begin{equation}
\label{eq:pstar_v2}
P^\star(c_i) =
\begin{cases}
\sigma\!\big(\beta\,\ell_{\mathrm{NLI}} + (1{-}\beta)\,\ell_{\mathrm{KG}}\big), \\
\quad \text{if } s_{\mathrm{KG}}\!<\!0.3 \text{ and } p_{\mathrm{NLI}}\!>\!0.7 \\[3pt]
p_{\mathrm{NLI}}(c_i), \quad \text{otherwise.}
\end{cases}
\end{equation}

% \begin{equation}
% \label{eq:pstar_v2}
% P^\star(c_i) = 
% \begin{cases}
% \sigma\!\big(\beta\,\mathrm{logit}(p_{\mathrm{NLI}}) \\
% \quad + (1{-}\beta)\,\mathrm{logit}(s_{\mathrm{KG}})\big), \\
% \quad\text{if } s_{\mathrm{KG}}(c_i)\!<\!0.3 \text{ and } 
% p_{\mathrm{NLI}}(c_i)\!>\!0.7 \\[2pt]
% p_{\mathrm{NLI}}(c_i), \;\; \text{otherwise}
% \end{cases}
% \end{equation}
Gating fires only when NLI is confident of entailment but KG is 
confident of conflict---the regime where a false KG signal would 
do the most damage. We tune $(\alpha, \beta, \tau)$ on dev 
(Appendix~\ref{app:calibration_details}).

\subsection{Metrics}
\label{subsec:metrics}

From $P^\star(c_i)$ and the predicted labels, we derive two core answer-level metrics:
\emph{Faithfulness} $\mathrm{Faith}(a) = \frac{1}{n}\sum_i \mathbb{I}[\hat y_i{=}\textsc{Entail}]$ and
\emph{Hallucination rate} $\mathrm{Halluc}(a) = \frac{1}{n}\sum_i \mathbb{I}[\hat y_i{=}\textsc{Contradict}]$.
We additionally compute (i) claim extraction quality (span-F1 and 
BERTScore~\citep{zhang2019bertscore}), (ii) retrieval diagnostics (ClaimRec, CtxPrec, 
answer-level ClaimF1), (iii) a parametric SelfKnow score (NLI-only, 
empty context), and (iv) a SafetyErr rate — the fraction of 
KG-aligned safety-critical claims (drug--disease, drug--adverse-event) 
labeled \textsc{Contradict}, isolating clinically dangerous errors 
from benign inaccuracies.
% We additionally compute claim extraction quality (span-F1 and BERTScore against teacher claims),
% retrieval diagnostics (ClaimRec, CtxPrec),
% answer-level ClaimF1,
% a parametric SelfKnow score (NLI-only, empty context),
% and a SafetyErr rate---the fraction of KG-aligned safety-critical claims
% (e.g., drug--disease, drug--adverse-event) labeled \textsc{Contradict}---that
% isolates clinically dangerous errors from benign inaccuracies.
Formal definitions of all metrics are in Appendix~\ref{app:metrics_full}.
A \emph{decision-flip} claim is one whose verdict (Entail vs.\ not-Entail) differs between NLI-only and the Fused setting; flips are the loci where KG augmentation actually changes evaluation outcomes.

\begin{table*}[t]
\centering
\scriptsize
\setlength{\tabcolsep}{4pt}
\renewcommand{\arraystretch}{1.0}
\begin{adjustbox}{width=\textwidth,center}
\begin{tabular}{@{} l cc | ccccc @{}}
\toprule
\textbf{Generator} &
\multicolumn{2}{c|}{\textbf{Panel A: Extractor}} &
\multicolumn{5}{c}{\textbf{Panel B: Checker (SFT)}} \\
\cmidrule(lr){2-3}\cmidrule(lr){4-8}
& \textbf{BERT-F1} $\uparrow$
& \textbf{\#Claims}
& \textbf{Acc} (\%) & \textbf{Mac-F1} (\%) & \textbf{F1-E} (\%) & \textbf{F1-N} (\%) & \textbf{F1-C} (\%) \\
\midrule
Med-Qwen2-7B    & 78.3 & 3.25
  & 83.7 & 55.0 & 40.1 & \cellcolor{hlgreen}94.9 & \cellcolor{hlgreen}\textbf{42.4} \\
Med42-Llama3-8B & \cellcolor{hlgreen}\textbf{83.6} & 4.70
  & \cellcolor{hlgreen}85.6 & \cellcolor{hlgreen}\textbf{67.3} & \cellcolor{hlgreen}50.3 & 94.5 & 27.3 \\
Meditron3-8B    & 82.1 & \textbf{5.28}
  & 81.2 & 49.7 & 46.3 & 92.9 & \cellcolor{hlred}23.4 \\
\midrule
\multicolumn{3}{l}{Ensemble (F1-wt)} & \textbf{87.4} & 60.5 & \textbf{52.8} & \textbf{95.0} & 41.3 \\
\bottomrule
\end{tabular}
\end{adjustbox}
\caption{Distilled claim extractor and checker fidelity to the 
GPT-4.1 teacher. 
\textbf{Panel~A:} BERTScore F1 between student-extracted and 
teacher-extracted claims, and average number of claims per answer. 
\textbf{Panel~B:} Checker accuracy and per-class F1 on the 3-way 
NLI task. F1-E, F1-N, and F1-C denote per-class F1 for the 
\textsc{Entail}, \textsc{Neutral}, and \textsc{Contradict} labels 
respectively. \textbf{Bold} marks the column-wise maximum 
(including the ensemble row where reported). 
\colorbox{hlgreen}{Green} highlights the best value \emph{among 
single students only}; \colorbox{hlred}{red} highlights the worst. 
The ensemble row is excluded from the green/red comparison.}
\label{tab:combined_results}
% \caption{Distilled claim extractor and checker fidelity to the 
% GPT-4.1 teacher. 
% \textbf{Panel~A:} BERTScore F1 between student-extracted and 
% teacher-extracted claims, and average number of claims per answer. 
% \textbf{Panel~B:} Checker accuracy and per-class F1 on the 3-way 
% NLI task. F1-E, F1-N, and F1-C denote per-class F1 for the 
% \textsc{Entail}, \textsc{Neutral}, and \textsc{Contradict} labels 
% respectively. \textbf{Bold} marks the best value in each column.
% \colorbox{hlgreen}{Green} highlights the best value across single 
% students; \colorbox{hlred}{red} highlights the worst.}
\label{tab:combined_results}
\end{table*}

\begin{table*}[!htbp]
\centering
\small
\setlength{\tabcolsep}{3.0pt}
\renewcommand{\arraystretch}{1.1}
\begin{adjustbox}{width=\textwidth,center}
\begin{tabular}{@{}llccc|ccc@{}}
\toprule
& & \multicolumn{3}{c|}{\textbf{Teacher-checker (GPT-4.1)}} & \multicolumn{3}{c}{\textbf{Student-checker (SFT ensemble)}} \\
\cmidrule(lr){3-5}\cmidrule(lr){6-8}
\textbf{Generator} & \textbf{KG Setting} &
\textbf{Faith.} $\uparrow$ & \textbf{Halluc.} $\downarrow$ & \textbf{SafetyErr} $\downarrow$ &
\textbf{Faith.} $\uparrow$ & \textbf{Halluc.} $\downarrow$ & \textbf{SafetyErr} $\downarrow$ \\
\midrule

\multirow{4}{*}{Med-Qwen2-7B}
& NLI-only       & 68.1 & 21.8 & 29.9 & 62.3  & 20.7  & 19.6 \\
& + DRKG         & \cellcolor{hlgreen}\textbf{75.6} & \cellcolor{hlgreen}\textbf{18.6} & \cellcolor{hlgreen}\textbf{24.4} & \cellcolor{hlgreen}68.6 & \cellcolor{hlgreen}10.4  & \cellcolor{hlgreen}\textbf{12.6} \\
& + BioPortal (vanilla) & \cellcolor{hlgreen}72.5 & \cellcolor{hlred}23.4 & \cellcolor{hlgreen}25.7 & \cellcolor{hlgreen}68.8 & \cellcolor{hlred}20.8 & \cellcolor{hlred}18.3 \\
& + BioPortal (gated)   & \cellcolor{hlgreen}72.1 & \cellcolor{hlred}22.3 & \cellcolor{hlgreen}25.4 & \cellcolor{hlgreen}\textbf{69.6}  & \cellcolor{hlgreen}\textbf{9.1}  & \cellcolor{hlgreen}16.8 \\
\midrule

\multirow{4}{*}{Med42-Llama3-8B}
& NLI-only       & 70.2 & 18.1 & 29.8 & 67.3  & 24.3  & 28.9 \\
& + DRKG         & \cellcolor{hlgreen}\textbf{77.0} & \cellcolor{hlred}23.0 & \cellcolor{hlgreen}17.5 & \cellcolor{hlred}58.4  & \cellcolor{hlred}25.7  & \cellcolor{hlgreen}\textbf{19.6} \\
& + BioPortal (vanilla) & \cellcolor{hlgreen}73.3 & \cellcolor{hlgreen}16.7 & \cellcolor{hlgreen}16.8 & \cellcolor{hlgreen}67.8 & \cellcolor{hlred}24.8 & \cellcolor{hlred}24.4 \\
& + BioPortal (gated)   & \cellcolor{hlgreen}73.2 & \cellcolor{hlgreen}\textbf{14.3} & \cellcolor{hlgreen}\textbf{16.4} & \cellcolor{hlgreen}\textbf{69.1}   & \cellcolor{hlgreen}\textbf{22.4}  & \cellcolor{hlgreen}23.7 \\
\midrule

\multirow{4}{*}{Meditron3-8B}
& NLI-only       & 58.2 & \textbf{20.6} & 41.8 & 52.7  & 14.6  & 29.0 \\
& + DRKG         & \cellcolor{hlgreen}66.8 & \cellcolor{hlred}33.2 & \cellcolor{hlgreen}\textbf{23.7} & \cellcolor{hlgreen}\textbf{59.6}  & \cellcolor{hlred}27.3  & \cellcolor{hlgreen}27.2 \\
& + BioPortal (vanilla) & \cellcolor{hlgreen}67.9 & \cellcolor{hlred}23.6 & \cellcolor{hlgreen}37.9 & \cellcolor{hlgreen}53.4 & \cellcolor{hlred}15.7 & \cellcolor{hlgreen}26.7 \\
& + BioPortal (gated)   & \cellcolor{hlgreen}\textbf{69.3} & \cellcolor{hlred}23.5 & \cellcolor{hlgreen}36.7 & \cellcolor{hlgreen}54.6  & \cellcolor{hlgreen}\textbf{13.5}  & \cellcolor{hlgreen}\textbf{25.7} \\

\bottomrule
\end{tabular}
\end{adjustbox}
\caption{\textbf{Coverage-driven verification behavior.}
KG fusion results per generator (macro-averaged across four datasets).
DRKG's limited coverage (62.7\%) inflates hallucination on uncovered
claims (e.g., Meditron3-8B: 14.6\% $\to$ 27.3\%) via false
contradictions, while BioPortal's 94.2\% coverage reverses this
pattern (Meditron3-8B: 14.6\% $\to$ 13.5\%) and consistently reduces
safety-critical errors across generators.
\textbf{Bold} marks the best value within each generator block 
and checker setting.
\colorbox{hlgreen}{Green}/\colorbox{hlred}{red} indicate
improvement/degradation vs.\ NLI-only.
``vanilla'' applies Eq.~\ref{eq:pstar_fusion} to all KG-linked 
claims; ``gated'' uses Eq.~\ref{eq:pstar_v2}.}
\label{tab:kg_fusion_combined}
\end{table*}

\section{Experiments}
\label{sec:experiments}

\subsection{Experimental Setup}
\label{sec:exp_setup}
\paragraph{Data and RAG setup.}
We evaluated on four biomedical QA benchmarks: PubMedQA~\cite{jin-etal-2019-pubmedqa}, MedQuAD~\cite{BenAbacha-BMC-2019}, LiveQA~\cite{yang2017cmu}, and MedRedQA~\cite{nguyen-etal-2023-medredqa}. We use a PubMed-based RAG pipeline (Appendix~\ref{app:retrieval}) with fixed retrieval settings across runs and compare three biomedical generators:  Meditron3-8B, Med-Qwen2-7B, and Med42-Llama3-8B. PMC-LLaMA-13B is excluded from the main comparison due to degenerate extraction behavior (only 1.19 claims per answer vs.\ 3--5 for the other generators, and BERT-F1=45.7 against teacher claims vs.\ 78--84), and is reported separately in Appendix~\ref{app:full_diagnostics}.
% PMC-LLaMA-13B is excluded from the main comparison due to degenerate extraction behavior — only 1.19 claims per answer (vs.\ 3--5 for the other generators) and BERT-F1=45.7 against teacher claims (vs.\ 78--84) — and is reported separately in Appendix~\ref{app:full_diagnostics}.

\paragraph{Checker training and evaluation protocol.}
We use GPT-4.1 to provide claims and NLI labels on train/dev, then distill (i) a student claim extractor and (ii) student NLI checkers. Unless stated otherwise, we use the F1-weighted ensemble (Section~\ref{sec:method_nli}) for downstream diagnostics. We report three verification settings that differ only in who performs extraction/checking: \textbf{teacher checker} (GPT-4.1), \textbf{single-student}, and \textbf{student ensemble}. We split the teacher-labeled data into train/dev/test (80/10/10) and fine-tuned open-source students with low-rank adaptation
(LoRA)~\citep{hu2022lora}; training details are in Appendix~\ref{app:student_configs}. For KG-enhanced verification, we computed a KG support score 
from each backend (DRKG or BioPortal, Section~\ref{sec:method_kg}) 
and fused it with textual NLI via Eq.~\ref{eq:pstar_fusion} 
or Eq.~\ref{eq:pstar_v2}; $(\alpha,\beta,\tau)$ were tuned on dev and then fixed for all test results. GRPO refinements were exploratory and are reported in Appendix~\ref{app:grpo}; unless noted, all test results use student models only (no GPT-4.1 calls at inference).

\paragraph{Computational cost.}
\label{sec:cost_analysis}
The student ensemble processes ${\sim}$180 claims/min on a single A100-80GB
(${\sim}$0.3\,s and \$0.0004 per claim), roughly 50$\times$ cheaper and
7$\times$ faster than the GPT-4.1 teacher (${\sim}$2.1\,s, \$0.02 per claim).
End-to-end evaluation of 100 questions takes ${\sim}$2.5 minutes
with the student pipeline versus ${\sim}$18 minutes with the teacher.

\subsection{KG Back-end Instantiations}
\label{sec:exp_kg_backends}

To study how KG choice affects verification, we instantiate the KG 
module with two back-ends that share the same linking/scoring/fusion 
interface but differ in scope and edge-attribute richness:

\noindent\textbf{DRKG}~\citep{drkg2020} is a drug-repurposing-focused 
biomedical KG containing $\approx$97K entities and $\approx$5.8M 
edges, primarily covering drug--gene--disease relations. Its edges 
carry minimal attribute information beyond the relation type, and 
we score plausibility using pre-trained TransE embeddings. On our 
benchmarks, DRKG covers 62.7\% of the entity pairs extracted from 
claims (averaged over the three main generators; PMC-LLaMA-13B 
excluded for consistency with §\ref{sec:exp_setup}).

\noindent\textbf{BioPortal}~\citep{salvadores2013bioportal} is a 
multi-ontology hub integrating biomedical ontologies (SNOMED-CT, 
MeSH, NCIt, DOID, RxNorm) with cross-ontology links, accessed via 
the public API in early 2026. Its edges carry richer ontology-level 
attributes (direct property relations such as 
\texttt{indication}/\texttt{contraindication}, hierarchical 
ancestors, LOOM mappings), which support ontology-based scoring 
without requiring trained embeddings. BioPortal covers 94.2\% of 
our claim entity pairs (a 31.5pp increase over DRKG).

The two back-ends differ both in \emph{coverage} over our claim 
distribution and in the \emph{type of edge attributes} available 
for scoring. Section~\ref{sec:rq_coverage} analyzes how these 
differences propagate to downstream verification quality.

\subsection{Research Questions}
 
We organize experiments around three questions: how KG coverage
affects verification quality (Sec.~\ref{sec:rq_coverage}), whether
selective gating mitigates low-coverage failures
(Sec.~\ref{sec:rq_gating}), and how diagnostics align with human
judgment (Sec.~\ref{sec:rq_human}). Claim extraction and checker
distillation quality are validated in
Appendix~\ref{app:full_diagnostics}.

\subsection{Distillation Quality}
\label{sec:sanity_check}
 
Before presenting our main results, we briefly validate that
the distilled student checkers and extractors are faithful enough
to serve as the front-end of our pipeline. A full analysis is in
Appendix~\ref{app:full_diagnostics}; here we summarize the key
findings.
 
\paragraph{Extractor.}
Student extractors achieve BERTScore F1 of $0.78$--$0.84$ against
teacher claims (Med42-Llama3-8B: 83.6; Meditron3-8B: 82.1;
Med-Qwen2-7B: 78.3), producing 3--5 atomic claims per answer that
closely match the teacher's decomposition. Qualitative analysis
(Appendix~\ref{app:extraction_errors}) further shows that
hallucinated claims are rare (6\%) and over-splitting is the
dominant but benign failure mode.
 
\paragraph{Checker ensemble.}
After SFT, the F1-weighted student ensemble reaches macro-F1 of
60.5 on the 3-way NLI task ( Table~\ref{tab:combined_results}),
close to the teacher's performance, and is more robust than any
single checker on the minority \textsc{Contradict} class.
 
These results establish that downstream variations we observe in
the main experiments are attributable to the verification
strategy (NLI-only vs.\ KG-augmented) rather than to poor
extraction or checker quality.

\begin{figure}[t]
\centering
\includegraphics[width=\columnwidth]{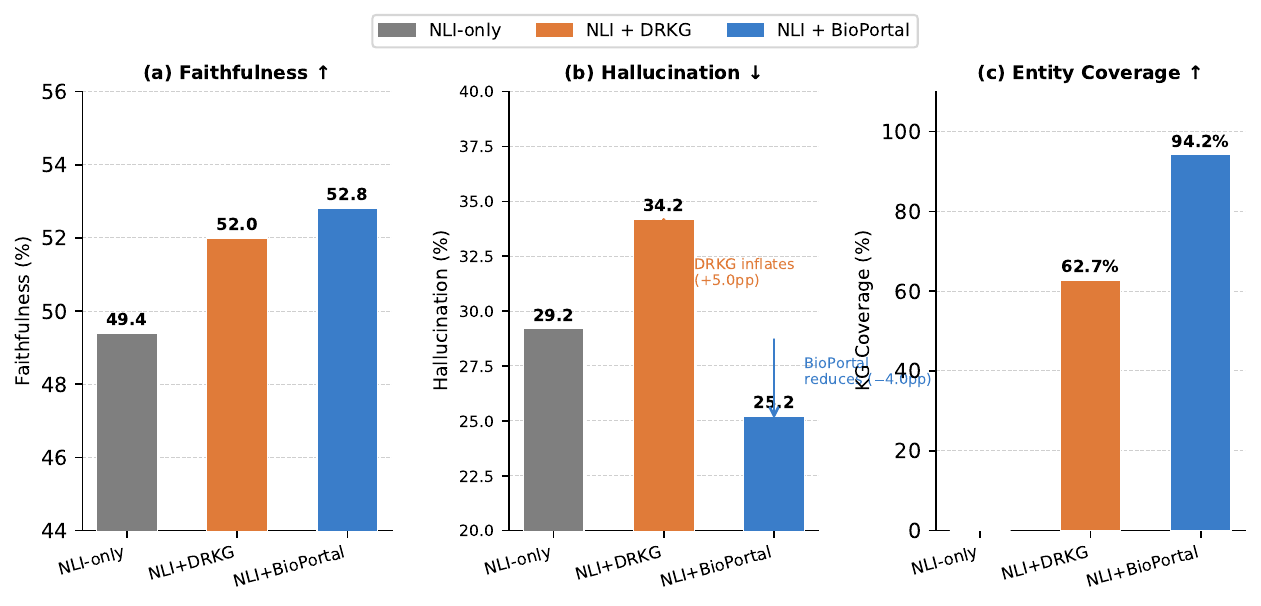}
\caption{Effect of KG back-end on verification quality. (a)~BioPortal improves faithfulness by +3.4pp over NLI-only. (b)~DRKG's limited coverage inflates hallucination to 34.2\%; BioPortal reduces it to 25.2\%. (c)~BioPortal achieves 94.2\% entity coverage vs.\ DRKG's 62.7\%.}
\label{fig:kg_comparison}
\end{figure}

\subsection{RQ1: KG Coverage Determines Verification Quality}
\label{sec:rq_coverage}

% To isolate the effect of the gating rule from KG coverage, we compare 
% selective gating (Eq.~\ref{eq:pstar_v2}) against a \emph{vanilla 
% fusion} baseline that applies Eq.~\ref{eq:pstar_fusion} to all 
% KG-linked claims without the gating condition. Both variants share 
% the same KG and fusion weights, so the comparison isolates the 
% gating rule itself.

We compare three verification settings under identical student
checkers: NLI-only, NLI$+$DRKG, and NLI$+$BioPortal.
Figure~\ref{fig:kg_comparison} summarizes the macro-averaged result;
per-generator breakdowns are in Table~\ref{tab:kg_fusion_combined}.
 
\paragraph{Main finding: coverage swings hallucination by 9.0pp.}
Both KGs improve faithfulness over NLI-only
($+$2.6pp for DRKG, $+$3.4pp for BioPortal), but their effect on
hallucination rate diverges sharply:
DRKG \emph{inflates} hallucination from 29.2\% to 34.2\%
($+$5.0pp), while BioPortal \emph{reduces} it to 25.2\%
($-$4.0pp). Since both back-ends share identical linking, scoring,
and fusion machinery and differ primarily in coverage (62.7\% vs.\
94.2\%) we attribute this 9.0pp swing to coverage
(cf.\ Section~\ref{sec:exp_kg_backends} for the three-part 
argument).
 
\paragraph{Per-generator consistency.}
Table~\ref{tab:kg_fusion_combined} shows the effect holds across
all three generators under the student-checker setting.
For Meditron3-8B, DRKG nearly doubles hallucination
(14.6\% $\to$ 27.3\%) while BioPortal keeps it flat
(14.6\% $\to$ 13.5\%). SafetyErr, in contrast, drops under
\emph{both} KGs, indicating that KG signals remain reliable
when they fire on safety-critical covered relations---
the failure mode is restricted to uncovered claims leaking weak
signals into the fusion.
 
\paragraph{Per-dataset coverage effect.}
Coverage also varies by dataset
(Appendix~\ref{app:kg_per_dataset}, Table~\ref{tab:kg_comparison_per_dataset}).
DRKG covers only 45\% of PubMedQA claims 
(Table~\ref{tab:kg_comparison_per_dataset}; research-oriented
content is underrepresented in a drug-repurposing KG), rising
to 78\% on LiveQA (drug-centric consumer questions).
BioPortal maintains $\geq$93\% coverage across all four
datasets. The hallucination-reduction gap between BioPortal
and DRKG is largest exactly where DRKG coverage is lowest
(PubMedQA, MedQuAD), further supporting the coverage-driven
interpretation. Per-generator coverage and KG-subset behavior are in Table~\ref{tab:kg_coverage}.

\paragraph{Human validation on decision-flip.}
On claims where NLI-only and Fused disagree, human annotators 
agree with BioPortal-fused predictions 69.8\% of the time 
vs.\ NLI-only's 63.4\% ($+$6.4pp; on 31\% of all flips, the KG-fused verdict matches the human label while NLI-only does not).
Macro-F1 improves from 59.2 to 64.7. Hallucination-rate 
changes thus reflect genuine accuracy gains, not metric 
artifacts. Full protocol is in Appendix~\ref{app:human_flip}.

\subsection{RQ2: Selective Gating Mitigates Low-Coverage Failure}
\label{sec:rq_gating}
 
The main result in RQ1 uses selective gating
(Eq.~\ref{eq:pstar_v2}) in the BioPortal configuration.
Here we ablate gating to isolate its contribution.

 \textbf{Vanilla fusion baseline.}
To separate the effect of the gating rule from the effect of KG 
coverage, we compare selective gating (Eq.~\ref{eq:pstar_v2}) 
against a \emph{vanilla fusion} baseline that applies 
Eq.~\ref{eq:pstar_fusion} to all KG-linked claims without the 
gating condition. Both variants share the same KG and fusion 
weights $(\alpha, \beta, \tau)$, so any difference in 
Table~\ref{tab:kg_fusion_combined} between BioPortal (vanilla) 
and BioPortal (gated) rows isolates the gating rule itself. 

\textbf{Gating trigger rate.}
The gating condition
($s_{\mathrm{KG}} < 0.3$ and $p_{\mathrm{NLI}} > 0.7$) fires on
12.3\% of KG-linked claims on average across our four datasets
(per-dataset range 9.4\%--14.7\%; see Appendix~\ref{app:kg_per_dataset}).
For the remaining 88\%, the KG signal is either weak or ambiguous,
and the fused score reduces to $p_{\mathrm{NLI}}$, preserving
NLI-only behavior.

\textbf{Gating ablation on the same KG.}
Comparing BioPortal (vanilla) and BioPortal (gated) in 
Table~\ref{tab:kg_fusion_combined} isolates gating's effect 
under fixed coverage: across all three generators, gating 
consistently reduces hallucination (by 0.7--2.4pp depending on 
generator and checker) while maintaining comparable faithfulness, 
confirming the gating rule, not just high coverage, avoids 
false contradictions.
We do not report a DRKG$+$gated variant because DRKG's TransE 
scores concentrate below the gating threshold, so gating fires 
indiscriminately and the gated variant behaves essentially as 
vanilla DRKG.

\textbf{Relation-type stratification.}
Appendix~\ref{app:kg_case} decomposes decision-flip outcomes by
relation type: KG correctly overrides NLI on 72\% of
drug--adverse-event flips and 68\% of drug--disease-treatment
flips, but \emph{hurts} 23\% of general lifestyle claims through
spurious hierarchical matches. Selective gating is designed
precisely to suppress this second category by keeping NLI-only
whenever the KG signal is not decisive. This stratified view
confirms that gating concentrates KG influence on relations
where the KG's structural priors are clinically meaningful.

\subsection{RQ3: Human Alignment on Safety-Critical Claims}
\label{sec:rq_human}
%  Our human evaluation comprises 100 questions $\times$ 3 generators 
% $\times$ 2 annotators, yielding 600 paired ratings per dimension, comparable in scale to prior claim-level human evaluations such 
% as FActScore~\citep{min2023factscore}.
% Our RQ3 focuses on two targeted human studies that directly
% test the validity of our framework:
% (i) a claim-extraction sanity check, and
% (ii) the decision-flip study already summarized in
% Figure~\ref{fig:kg_comparison}(c).
Our RQ3 reports two targeted human studies on 100 questions $\times$ 3 generators $\times$ 2 annotators (comparable to FActScore~\citep{min2023factscore}): (i) a claim-extraction sanity check, and (ii) the decision-flip study summarized in Figure~\ref{fig:kg_comparison}(c).
% We deliberately avoid aggregate correctness/completeness
% ratings of long-form medical answers, which which in our pilot showed low inter-annotator agreement
% (quadratic-weighted Cohen's $\kappa = 0.32$ for correctness,
% $\kappa = 0.27$ for completeness;
% comparable to Med-PaLM physician ratings~\citep{singhal2025toward}).
% Such low agreement reflects the inherent subjectivity of rating
% open-ended medical answers holistically and does not speak to
% claim-level verification, which is our contribution.
% Full results are in Appendix~\ref{app:human_stats}.
 We deliberately avoid aggregate correctness/completeness
ratings of long-form medical answers, which in our pilot
showed low inter-annotator agreement (consistent with the
known difficulty of holistic medical answer rating, e.g.,
Med-PaLM physician ratings~\citep{singhal2025toward}).
Such low agreement reflects the inherent subjectivity of
rating open-ended medical answers holistically and does
not speak to claim-level verification, which is our
contribution. Full agreement statistics are in
Appendix~\ref{app:human_stats}.
% \paragraph{Extractor quality (sanity check).}
% Inter-annotator agreement on claim-set quality is substantial
% ($\kappa = 0.67$), and all three generators achieve near-ceiling
% scores ($\geq 4.62/5$; Table~\ref{tab:human_ratings}).
% This confirms that the distilled extractor produces faithful
% atomic decompositions suitable for downstream claim-level
% verification.
 
% \paragraph{Extractor quality (sanity check).}
% Inter-annotator agreement on claim-set quality is substantial
% ($\kappa = 0.67$), and all three generators achieve near-ceiling
% scores ($\geq 4.62/5$; Table~\ref{tab:human_ratings}).
% This confirms that the distilled extractor produces faithful
% atomic decompositions suitable for downstream claim-level
% verification.
\textbf{Human ratings of extracted claim quality.}
Med-Qwen2-7B: $4.97 \pm 0.18$;
Med42-Llama3-8B: $4.73 \pm 0.99$;
Meditron3-8B: $4.62 \pm 0.18$
(1--5 Likert; two annotators, 100 questions; $\kappa = 0.67$).
Near-ceiling scores confirm the distilled extractor produces
faithful atomic decompositions. Aggregate answer-level
ratings (correctness, completeness) showed lower agreement
and are deferred to Appendix~\ref{app:human_stats}.
% \textbf{Human ratings of extracted claim quality.}
% Med-Qwen2-7B: $4.97 \pm 0.18$;
% Med42-Llama3-8B: $4.73 \pm 0.99$;
% Meditron3-8B: $4.62 \pm 0.18$
% (1--5 Likert; two annotators, 100 questions; $\kappa = 0.67$).
% Near-ceiling scores confirm the distilled extractor produces
% faithful atomic decompositions. Answer-level ratings (correctness,
% completeness; $\kappa < 0.35$) are in Appendix~\ref{app:human_stats}.
% \paragraph{Decision-flip agreement.}
% As reported in RQ1, BioPortal$+$gating improves human agreement
% on decision-flip cases from 63.4\% (NLI-only) to 69.8\%,
% with consistent improvements on safety-critical
% drug--disease and drug--adverse-event subsets
% (Appendix~\ref{app:kg_case}). This is our primary human-validated
% evidence that coverage-aware KG fusion yields better verdicts,
% not just different ones.
\textbf{Decision-flip agreement.}
RQ1 already established that BioPortal$+$gating raises human agreement on decision-flip cases by $+6.4$pp, with consistent improvements on safety-critical drug--disease and drug--adverse-event subsets (Appendix~\ref{app:kg_case}).
\subsection{Comparison with Existing RAG Evaluators}
\label{sec:comparison}
 
To position \textsc{MedRAGChecker} against general-purpose RAG evaluation
tools, we compare with RAGAS~\citep{es-etal-2024-ragas} and
RAGChecker~\citep{ru2024ragchecker} on answer-level faithfulness
using the same retrieval pipeline and generator outputs
(Table~\ref{tab:baseline_comparison}).
We use Med42-Llama3-8B as a representative generator (best macro-F1 as a checker; comparable claim quality across generators, see \S\ref{sec:rq_human}).
% We use Med42-Llama3-8B as a representative generator (best macro-F1 as a checker; comparable claim quality in Table~\ref{tab:human_ratings}).

\begin{table}[t]
\centering
\small
\setlength{\tabcolsep}{2.5pt}
\renewcommand{\arraystretch}{1.05}
\begin{tabular}{llccc}
\toprule
& &
\textbf{MedRAG-} & & \textbf{RAG-} \\
\textbf{Dataset} & \textbf{Metric} &
\textbf{Checker} & \textbf{RAGAS} & \textbf{Checker} \\
\midrule
\multirow{2}{*}{PubMedQA}
 & Faith. $\uparrow$    & 88.9 & \textbf{89.5} & 87.2 \\
 & Halluc. $\downarrow$ & \textbf{4.7}  & \NA  & 5.1 \\
\midrule
\multirow{2}{*}{LiveQA}
 & Faith. $\uparrow$    & \textbf{82.5} & 74.9 & 70.8 \\
 & Halluc. $\downarrow$ & \textbf{8.6}  & \NA  & 19.5 \\
\midrule
\multirow{2}{*}{MedQuAD}
 & Faith. $\uparrow$    & \textbf{87.6} & 84.7 & 83.9 \\
 & Halluc. $\downarrow$ & \textbf{5.5}  & \NA  & 6.8 \\
\midrule
\multirow{2}{*}{MedRedQA}
 & Faith. $\uparrow$    & \textbf{82.4} & 82.1 & 81.5 \\
 & Halluc. $\downarrow$ & \textbf{6.4}  & \NA  & 8.2 \\
\bottomrule
\end{tabular}
\caption{Comparison with general-purpose RAG evaluators.
All numbers use Med42-Llama3-8B as a representative generator
to enable controlled comparison on the same outputs.
\NA{} indicates the metric is not supported by that tool.
\textbf{Bold} marks the best value in each row.
\textsc{MedRAGChecker} achieves comparable faithfulness while
additionally providing claim-level diagnostics, biomedical KG
verification, and safety-critical error detection.}
\label{tab:baseline_comparison}
\end{table} 
% \begin{table}[t]
% \centering
% \small
% \setlength{\tabcolsep}{2.5pt}
% \renewcommand{\arraystretch}{1.05}
% \begin{tabular}{llccc}
% \toprule
% & &
% \textbf{MedRAG-} & & \textbf{RAG-} \\
% \textbf{Dataset} & \textbf{Metric} &
% \textbf{Checker} & \textbf{RAGAS} & \textbf{Checker} \\
% \midrule
% \multirow{2}{*}{PubMedQA}
%  & Faith. $\uparrow$    & 88.9 & 89.5          & 87.2 \\
%  & Halluc. $\downarrow$  & 4.7  & \NA           & 5.1 \\
% \midrule
% \multirow{2}{*}{LiveQA}
%  & Faith. $\uparrow$    & 82.5 & 74.9          & 70.8 \\
%  & Halluc. $\downarrow$  & 8.6  & \NA           & 19.5 \\
% \midrule
% \multirow{2}{*}{MedQuAD}
%  & Faith. $\uparrow$    & 87.6 & 84.7   & 83.9 \\
%  & Halluc. $\downarrow$  & 5.5  & \NA           & 6.8 \\
% \midrule
% \multirow{2}{*}{MedRedQA}
%  & Faith. $\uparrow$    & 82.4 & 82.1   & 81.5 \\
%  & Halluc. $\downarrow$  & 6.4  & \NA           & 8.2 \\
% \bottomrule
% \end{tabular}
% \caption{Comparison with general-purpose RAG evaluators.
% All numbers use Med42-Llama3-8B as a representative generator
% to enable controlled comparison on the same outputs.
% \NA{} indicates the metric is not supported by that tool.
% \textsc{MedRAGChecker} achieves comparable faithfulness while
% additionally providing claim-level diagnostics, biomedical KG
% verification, and safety-critical error detection.}
% \label{tab:baseline_comparison}
% \end{table}

All three frameworks achieve comparable faithfulness, as expected from shared NLI-style cores.
\textsc{MedRAGChecker} differs in four respects: claim-level granularity; KG verification that improves human agreement on decision-flip (69.8\% vs.\ 63.4\%); safety-critical error rates unavailable in general-purpose tools; and API-free deployment via distillation.
% \section{Conclusion}
% \textsc{MedRAGChecker} provides a claim-level diagnostic framework 
% for biomedical RAG that combines distilled NLI checkers with 
% KG-based consistency under a coverage-aware selective gating rule. 
% It matches general-purpose RAG evaluators on faithfulness while 
% additionally surfacing safety-critical errors that align with 
% human judgment on decision-flip claims. Our supplementary analysis 
% with two KG back-ends shows that KG coverage matters substantially 
% in practice: an incomplete KG can inflate hallucination via false 
% contradictions, motivating the selective gating design. 
% Practitioners should select KG back-ends by coverage over the 
% target claim distribution and combine KG with textual signals 
% through gating rather than unconditional fusion. The methodology 
% should transfer to other high-stakes verification domains such as 
% legal and financial QA.
\section{Conclusion}
\textsc{MedRAGChecker} provides a claim-level diagnostic framework 
for biomedical RAG that matches general-purpose evaluators on 
faithfulness while surfacing safety-critical errors aligned with 
human judgment on decision-flip claims. Our two-back-end analysis 
further shows that KG coverage substantially affects fusion quality, 
motivating the selective gating design. The methodology should 
transfer to other high-stakes verification domains such as legal 
and financial QA.
\section*{Limitations}
\textbf{Teacher supervision as pseudo-ground truth.}
Our distillation and evaluation rely on teacher labels for claim 
decomposition and NLI judgments, which may contain biases or 
systematic errors, especially for rare biomedical conditions or 
ambiguous questions. Our teacher-sensitivity analysis 
(Appendix~\ref{app:teacher_sensitivity}, 
Table~\ref{tab:teacher_sensitivity}) confirms that generator 
rankings are stable under both GPT-4.1 and GPT-4o teachers 
across research-style datasets, though absolute values on 
consumer-health data (MedRedQA, $\kappa = 0.13$ between teachers) 
remain teacher-conditional.

\textbf{Checker calibration and class imbalance.}
Contradiction is typically a minority class. Even with ensembling, performance varies across datasets, and subtle contradictions can be under-detected. MedRedQA exhibits low inter-teacher agreement (Appendix Table~\ref{tab:teacher_sensitivity}); we therefore interpret absolute Faith/Halluc values on MedRedQA with caution and focus on within-teacher relative comparisons.  
% MedRedQA exhibits low inter-teacher agreement (Appendix Table~\ref{tab:teacher_sensitivity}), so we interpret absolute Faith/Halluc values on MedRedQA with caution and primarily focus on within-teacher relative comparisons and trends.

\textbf{KG coverage and linker errors.}

Our DRKG vs.\ BioPortal comparison shows that KG coverage substantially affects fusion quality (Section~\ref{sec:rq_coverage}).
Even BioPortal does not cover all biomedical
entities and relations. Entity linking relies on SapBERT-based
matching and BioPortal API search, which can fail for rare entities,
complex paraphrases, or multi-hop claims.
BioPortal's lower average consistency on some datasets
(Table~\ref{tab:kg_comparison_per_dataset}) reflects its
inclusion of weaker hierarchical signals;
our selective gating strategy (Eq.~\ref{eq:pstar_v2}) mitigates
this but does not fully resolve it.
For safety-critical relation families (drug--disease, drug--adverse-event), KG-based signals are most reliable; elsewhere, they complement but do not replace textual NLI verification.
General-purpose LLM-based evaluators may be less reliable in high-stakes medical settings without domain calibration, and standardized evaluation remains an open challenge for medical LLM applications.

\textbf{Circularity of teacher supervision.}
While the KG component is teacher-independent,
claim extraction and the primary NLI signal remain derived from
GPT-4.1, creating a partially circular evaluation loop.
A fully teacher-free pipeline would require human-annotated claims
at scale; we partially mitigate this through human validation
(Section~\ref{sec:rq_human}) and teacher-robustness analysis
(Appendix~\ref{app:teacher_sensitivity}).

\textbf{Domain generalization.}
Our methodology (claim extraction, NLI + KG verification, fusion)
transfers to other technical domains such as legal and financial QA,
but requires a domain-specific KG and re-distillation of student models.
Cross-domain evaluation is left for future work.

\textbf{Scale of human evaluation.}
Our human validation comprises 100 questions $\times$ 3 generators 
$\times$ 2 annotators, a scale comparable to prior claim-level 
human studies such as FActScore~\citep{min2023factscore}. 
Inter-annotator agreement is substantial on claim-set quality 
($\kappa = 0.67$) but lower on holistic correctness ratings, 
reflecting known difficulty in evaluating long-form medical 
answers. Decision-flip-level conclusions should
be interpreted accordingly. Larger-scale validation with
multiple clinician annotators is a natural next step.

% Inter-annotator agreement is substantial on claim-set quality 
% ($\kappa = 0.67$) but lower on holistic correctness ratings
% ($\kappa = 0.32$), reflecting known difficulty in evaluating 
% long-form medical answers. 

\section*{Ethics Statement}
\textsc{MedRAGChecker} is designed to \emph{detect} unsupported or contradictory claims in biomedical RAG outputs and
to surface safety-critical error patterns. Nevertheless, automatic verification can be wrong; false negatives may
miss harmful claims and false positives may over-warn. We recommend using \textsc{MedRAGChecker} as a screening and
debugging aid, with human oversight for high-stakes settings.

Our experiments use publicly available datasets and retrieved biomedical literature.

For our human evaluation, annotators were informed of the study  purpose, compensated appropriately, and instructed not to  provide medical advice.  We do not process private patient data. We will release prompts, model checkpoints (where licenses allow), and evaluation code to support transparency and reproducibility.

\section*{Acknowledgments}

\paragraph{Use of Generative AI.}
GPT-4.1 is used as a methodological component of this work,
serving as the teacher model for claim extraction
(Section~\ref{sec:method_extract}) and NLI verification
(Section~\ref{sec:method_nli}) in our distillation pipeline,
and as the basis for the teacher-sensitivity comparison
against GPT-4o (Appendix~\ref{app:teacher_sensitivity}).
The KG verification module operates without GPT-4.1 calls.
At inference time, we use only distilled student models;
the full mapping of models to pipeline stages is documented
in Section~\ref{sec:framework}. We also used general-purpose
LLM assistants for light editing and polishing of prose;
all technical content, experimental design, results, and
analyses are the authors' own.

\bibliography{custom}

\appendix

\section{Full Metric Definitions}
\label{app:metrics_full}

\paragraph{Notation.}
For an example $(q, D, a)$, the extractor outputs claims
$\mathcal{C}=\{c_i\}_{i=1}^n$.
The checker predicts $\hat y_i\in\{\textsc{Entail},\textsc{Neutral},\textsc{Contradict}\}$ and a fused confidence $P^\star(c_i)\in[0,1]$ (Eq.~\ref{eq:pstar_fusion}).
In the fused setting, we replace the textual NLI \textsc{Entail} logit with the fused logit from $P^\star(c_i)$, keep \textsc{Neutral}/\textsc{Contradict} logits unchanged, renormalize via softmax, and predict $\hat y_i=\arg\max_y p(y\mid c_i,D)$.
Binary supported/unsupported uses threshold $\tau$ (tuned on dev): $\mathbb{I}_{\text{sup}}(c_i)=\mathbb{I}[P^\star(c_i)\ge \tau]$.

\paragraph{Claim extraction quality.}
Against teacher claims (GPT-4.1), we compute soft span-level P/R/F1 via token-overlap matching (Appendix~\ref{app:span_f1}) and report average claims per answer.

\paragraph{Retrieval diagnostics.}
\begin{equation}
\mathrm{ClaimRec} = \frac{1}{|\mathcal{C}^{\text{ref}}|}
\sum_{c\in\mathcal{C}^{\text{ref}}} \mathbb{I}_{\text{sup}}(c),
\end{equation}
where $\mathcal{C}^{\text{ref}}$ are reference claims (reported only when references are available).
\begin{equation}
\mathrm{CtxPrec}
=\frac{1}{k}\sum_{j=1}^{k}
\mathbb{I}\Big[\exists i:\ d_j \text{ cited for } c_i
\wedge P^\star(c_i)\ge\tau\Big].
\end{equation}

\paragraph{ClaimF1.}
We match generated claims $\mathcal{C}^{\text{gen}}$ to reference claims $\mathcal{C}^{\text{ref}}$ by token-overlap, count a match as correct if supported ($P^\star\ge\tau$), and report the harmonic mean of Precision and Recall.

\paragraph{Self-knowledge.}
We disable KG fusion and rerun NLI with empty context $\varnothing$:
$\mathrm{SelfKnow} = \frac{1}{n}\sum_i \mathbb{I}[p_{\mathrm{NLI}}(c_i\mid\varnothing)\ge\tau_{\textsc{NLI}}]$.

\paragraph{Safety-critical error rate.}
$\mathrm{SafetyErr} = \frac{1}{|\mathcal{C}_{\text{safety}}|}\sum_{c\in\mathcal{C}_{\text{safety}}} \mathbb{I}[\hat y(c)=\textsc{Contradict}]$,
where $\mathcal{C}_{\text{safety}}$ are claims mapped to safety-critical KG relations.

\section{Dataset}
\label{app:dataset_details}
This section summarizes the four biomedical QA benchmarks used in our experiments.
Table~\ref{tab:dataset_stats} reports length and retrieval-context statistics,
and Table~\ref{tab:dataset_overview_qual} provides a qualitative overview of question style and evidence source.

\begin{table*}[t]
  \centering
  \small
  \setlength{\tabcolsep}{3pt} 
  \begin{tabular}{lrrrrrrrr}
    \hline
    \textbf{Dataset} &
    \textbf{\#Q} &
    \textbf{Median $|q|$} &
    \textbf{Max $|q|$} &
    \textbf{Median $|a|$} &
    \textbf{Max $|a|$} &
    \textbf{Median \#Doc} &
    \textbf{Max \#Doc} &
    \textbf{Median/Max $|d|$} \\
    \hline
     MedRedQA & 799  & 167 & 3732 & 56 & 1187 & 8  & 8  & 123 / 187 \\
    TREC LiveQA Medical        & 104  & 12  & 66   & 88 & 486  & 8  & 8  & 205 / 972 \\
    MedQuAD       & 1000 & 9   & 18   & 86 & 731  & 10 & 10 & 123 / 203 \\
    PubMedQA      & 1000 & 15  & 43   & 35 & 263  & 8  & 8  & 239 / 735 \\
    \hline
  \end{tabular}
  \caption{Dataset statistics for MedRAGChecker RAG inputs. $|q|$ and $|a|$ denote tokenized question and (ground-truth) answer lengths (whitespace tokens). $|d|$ denotes retrieved document length; median/max refer to median and maximum values across all retrieved contexts.}
  \label{tab:dataset_stats}
\end{table*}

\begin{table*}[t]
  \centering
  \small
  \begin{tabularx}{\linewidth}{p{2.3cm} p{3cm} p{3cm} p{6cm}}
    \hline
    \textbf{Dataset} &
    \textbf{Domain} &
    \textbf{Answer type} &
    \textbf{Question style / evidence} \\
    \hline
    % MedRedQA  \citep{nguyen-etal-2023-medredqa} &
    MedRedQA  &
    Consumer health (Reddit \texttt{r/AskDocs}) &
    Free-text answers by verified clinicians &
    Medium--long layperson questions covering diverse symptoms and concerns; thread content and PubMed-style references. \\
    \hline
    % TREC LiveQA Medical (2017) \citep{yang2017cmu} &
    TREC LiveQA Medical &
    Consumer health questions to NLM &
    Free-text answers with quality judgments &
    Long, noisy real-world user questions; reference answers plus retrieved passages with human relevance labels. \\
    \hline
    % MedQuAD \citep{BenAbacha-BMC-2019} &
    MedQuAD &
    NIH consumer health websites &
    Free-text factoid / definition QA &
    Short--medium consumer questions about diseases, drugs, and procedures; long snippets from NIH pages with source URLs. \\
    \hline
    % PubMedQA \citep{jin-etal-2019-pubmedqa} &
    PubMedQA &
    Biomedical research (PubMed abstracts) &
    Yes/No/Maybe + long-answer rationale &
    Short research questions derived from article titles; abstract body used as context, conclusion paragraph as gold answer. \\
    \hline
  \end{tabularx}
  \caption{Qualitative overview of the biomedical QA datasets used with MedRAGChecker, including domain, answer type, and typical question/evidence style.}
  \label{tab:dataset_overview_qual}
\end{table*}

\FloatBarrier
\section{Full Diagnostic Tables}
\label{app:full_diagnostics}

Per-generator end-to-end diagnostics are in
Table~\ref{tab:end_to_end_generators}; teacher-checker results
across both GPT-4.1 and GPT-4o are in
Table~\ref{tab:teacher_checker_full}; per-claim NLI label
distributions are in Table~\ref{tab:text_eval_summary}.

\begin{table*}[t]
\centering
\small
\tight
\begin{tabular}{llccccc}
\toprule
\textbf{Dataset} & \textbf{Generator} &
\textbf{Faith.} $\uparrow$ &
\textbf{Halluc.} $\downarrow$ &
\textbf{CtxPrec} $\uparrow$ &
\textbf{SelfKnow.} $\uparrow$ &
\textbf{SafetyErr} $\downarrow$ \\
\midrule
\multirow{4}{*}{PubMedQA}
& Med-Qwen2-7B       & 83.6 & 7.4 & 41.8 & 28.2 & 7.0 \\
& Med42-Llama3-8B    & \textbf{88.9} & 4.7 & 41.4 & \textbf{32.7} & 6.5 \\
& Meditron3-8B       & 69.2 & \textbf{3.3} & \textbf{53.2} & 30.4 & \textbf{5.9} \\
& PMC-LLaMA-13B      & 57.9 & 11.7 & 35.6 & 21.6 & 12.4 \\
\midrule
\multirow{4}{*}{MedQuAD}
& Med-Qwen2-7B       & 85.3 & 6.0 & 42.4 & \textbf{32.4} & 8.5 \\
& Med42-Llama3-8B    & \textbf{87.6} & \textbf{5.5} & 46.3 & 31.2 & \textbf{7.8} \\
& Meditron3-8B       & 73.1 & 7.6 & \textbf{49.4} & 24.0 & 8.2 \\
& PMC-LLaMA-13B      & 63.2 & 8.7 & 44.6 & 20.0 & 9.7 \\
\midrule
\multirow{4}{*}{LiveQA}
& Med-Qwen2-7B       & 72.4 & 11.3 & 43.2 & 23.1 & 7.6 \\
& Med42-Llama3-8B    & \textbf{82.5} & \textbf{8.6} & 42.7 & 28.6 & \textbf{6.2} \\
& Meditron3-8B       & 72.3 & 11.3 & \textbf{48.6} & \textbf{31.4} & 9.5 \\
& PMC-LLaMA-13B      & 55.7 & 12.8 & 38.1 & 22.6 & 10.4 \\
\midrule
\multirow{4}{*}{MedRedQA}
& Med-Qwen2-7B       & \textbf{84.2} & 7.3 & 41.7 & 26.4 & 7.6 \\
& Med42-Llama3-8B    & 82.4 & \textbf{6.4} & 42.7 & 27.2 & \textbf{6.7} \\
& Meditron3-8B       & 71.2 & 8.3 & \textbf{44.3} & \textbf{28.0} & 9.3 \\
& PMC-LLaMA-13B      & 63.4 & 9.5 & 41.3 & 22.5 & 12.7 \\
\bottomrule
\end{tabular}
\caption{Full end-to-end \textsc{MedRAGChecker} diagnostics across 
all datasets and generators (teacher = GPT-4.1). \textbf{Bold} marks 
the best value per dataset and column.}
\label{tab:end_to_end_generators}
\end{table*}
% \begin{table*}[t]
% \centering
% \small
% \tight
% \begin{tabular}{llccccc}
% \toprule
% \textbf{Dataset} & \textbf{Generator} &
% \textbf{Faith.} $\uparrow$ &
% \textbf{Halluc.} $\downarrow$ &
% \textbf{CtxPrec} $\uparrow$ &
% \textbf{SelfKnow.} $\uparrow$ &
% \textbf{SafetyErr} $\downarrow$ \\
% \midrule
% \multirow{4}{*}{PubMedQA}
% & Med-Qwen2-7B       & 83.6 & 7.4 & 41.8 & 28.2 & 7.0 \\
% & Med42-Llama3-8B    & 88.9 & 4.7 & 41.4 & 32.7 & 6.5 \\
% & Meditron3-8B       & 69.2 & 3.3 & 53.2 & 30.4 & 5.9 \\
% & PMC-LLaMA-13B      & 57.9 & 11.7 & 35.6 & 21.6 & 12.4 \\
% \midrule
% \multirow{4}{*}{MedQuAD}
% & Med-Qwen2-7B       & 85.3 & 6.0 & 42.4 & 32.4 & 8.5 \\
% & Med42-Llama3-8B    & 87.6 & 5.5 & 46.3 & 31.2 & 7.8 \\
% & Meditron3-8B       & 73.1 & 7.6 & 49.4 & 24.0 & 8.2 \\
% & PMC-LLaMA-13B      & 63.2 & 8.7 & 44.6 & 20.0 & 9.7 \\
% \midrule
% \multirow{4}{*}{LiveQA}
% & Med-Qwen2-7B       & 72.4 & 11.3 & 43.2 & 23.1 & 7.6\\
% & Med42-Llama3-8B    & 82.5 & 8.6 & 42.7 & 28.6 & 6.2 \\
% & Meditron3-8B       & 72.3 & 11.3 & 48.6 & 31.4 & 9.5 \\
% & PMC-LLaMA-13B      & 55.7 & 12.8 & 38.1 & 22.6 & 10.4 \\
% \midrule
% \multirow{4}{*}{MedRedQA}
% & Med-Qwen2-7B       & 84.2 & 7.3 & 41.7 & 26.4 & 7.6 \\
% & Med42-Llama3-8B    & 82.4 & 6.4 & 42.7 & 27.2 & 6.7 \\
% & Meditron3-8B       & 71.2 & 8.3 & 44.3 & 28.0 & 9.3 \\
% & PMC-LLaMA-13B      & 63.4 & 9.5 & 41.3 & 22.5 & 12.7 \\
% \bottomrule
% \end{tabular}

% \caption{RQ3: Full end-to-end MedRAGChecker diagnostics across all datasets and generators (teacher = GPT-4.1 ).}

% \label{tab:end_to_end_generators}
% \end{table*}
\begin{table*}[t]
\centering
\scriptsize
\setlength{\tabcolsep}{2.0pt}
\renewcommand{\arraystretch}{1.08}
\begin{adjustbox}{max width=\textwidth}
\begin{tabular}{@{} >{\raggedright\arraybackslash}p{1.6cm} l
r r r r r
r r r r r @{}}
\toprule
& & \multicolumn{5}{c}{Teacher = GPT-4.1} & \multicolumn{5}{c}{Teacher = GPT-4o} \\
\cmidrule(lr){3-7}\cmidrule(lr){8-12}
Dataset & Generator
& ClaimF1 & ClaimRec & CtxPrec & Faith. & Halluc.
& ClaimF1 & ClaimRec & CtxPrec & Faith. & Halluc. \\
\midrule

\multirow{4}{*}{PubMedQA}
& Med-Qwen2-7B        & 26.7 & 97.6 & 49.9 & 88.6 & 7.4  & \textbf{23.2} & 96.9 & 41.8 & 83.6 & 7.4  \\
& Med42-Llama3-8B     & \textbf{27.6} & 97.9 & 49.8 & \textbf{92.4} & 5.2  & 21.1 & \textbf{99.0} & 41.4 & \textbf{88.9} & 4.7  \\
& Meditron3-8B        & 23.5 & 97.2 & 32.9 & 69.2 & \textbf{3.3}  & 21.0 & 97.5 & 41.9 & 63.7  & \textbf{1.0}   \\
& PMC-LLaMA-13B       & 14.6 & \textbf{98.3} & \textbf{50.0} & 58.0 & 29.6 & 14.7 & 97.4 & \textbf{50.7} & 60.5 & 29.5 \\
\midrule

\multirow{4}{*}{MedQuAD}
& Med-Qwen2-7B        & 39.3 & \textbf{65.7} & 51.4 & \textbf{88.1} & 9.4  & 30.1 & 62.3 & 47.0 & \textbf{82.9} & \textbf{6.2}  \\
& Med42-Llama3-8B     & \textbf{43.4} & 61.5 & 51.4 & 86.1 & 6.4  & \textbf{39.5} & 62.8 & 46.0 & 80.2 & 8.5  \\
& Meditron3-8B        & 31.8 & 63.3 & \textbf{52.8} & 83.1 & \textbf{4.9}  & 30.1 & \textbf{63.9} & \textbf{49.4} & 73.1 & 7.6  \\
& PMC-LLaMA-13B       & 0.0  & 63.5 & 51.9 & 0.0  & 5.2  & 0.0  & 61.3 & 47.2 & 4.0  & 6.0  \\
\midrule

\multirow{4}{*}{LiveQA}
& Med-Qwen2-7B        & 6.3  & 26.8 & \textbf{43.1} & \textbf{72.4} & 18.3 & 5.3  & 23.8 & 37.6 & \textbf{65.9} & 22.4 \\
& Med42-Llama3-8B     & \textbf{10.8} & \textbf{27.5} & 42.2 & 70.3 & \textbf{12.6} & \textbf{8.8}  & 24.8 & 38.1 & 63.7 & \textbf{7.5}  \\
& Meditron3-8B        & 5.7  & 26.4 & 41.1 & 53.9 & 22.9 & 4.6  & \textbf{26.3} & \textbf{41.9} & 29.6 & 13.5 \\
& PMC-LLaMA-13B       & 0.0  & 30.1 & 45.6 & 27.3 & 45.2 & 0.0  & 15.6 & 27.5 & 0.0  & 0.0  \\
\midrule

\multirow{4}{*}{MedRedQA}
& Med-Qwen2-7B        & 10.7 & \textbf{16.2} & \textbf{12.3} & 31.4 & \textbf{48.0} & 1.5  & \textbf{13.9} & 9.5  & 12.7  & 5.6   \\
& Med42-Llama3-8B     & \textbf{11.5} & 14.6 & 11.2 & \textbf{32.0} & 48.0 & \textbf{2.0}  & 13.2 & 9.8  & \textbf{14.0}  & \textbf{4.5}   \\
& Meditron3-8B        & 7.4  & 14.5 & 11.3 & 26.7 & 51.4 & 1.2  & 13.5 & \textbf{10.1} & 9.7  & 6.0  \\
& PMC-LLaMA-13B       & 9.8  & 14.4 & 10.5 & 7.4  & 75.0 & 0.0  & 9.2  & 7.5  & 3.7  & 5.3  \\
\bottomrule
\end{tabular}
\end{adjustbox}
\caption{Teacher checker results (teacher = GPT-4.1 vs GPT-4o). 
\textbf{Bold} marks the best value per dataset, teacher, and column 
(excluding degenerate rows where Faith.\ or ClaimF1 collapses to 0).}
\label{tab:teacher_checker_full}
\end{table*}

\begin{table*}[t]
\centering
\small
\setlength{\tabcolsep}{4pt}
\renewcommand{\arraystretch}{1.10}
\begin{tabular}{l l r r r r r}
\toprule
Dataset & Generator & Avg.\ Resp.\ Claims & Avg.\ GT Claims & Ent (\%) & Neu (\%) & Con (\%) \\
\midrule

\multirow{4}{*}{PubMedQA}
& Med-Qwen2-7B        & 2.10 & 4.90  & 88.6 & 4.0  & 7.4 \\
& Med42-Llama3-8B     & 2.00 & 4.90  & 92.3 & 2.4  & 5.2 \\
& Meditron3-8B        & 1.95 & 4.90  & 29.1 & 69.0 & 1.8 \\
& PMC-LLaMA-13B       & 6.80 & 4.90  & 58.0 & 12.4 & 29.6 \\
\midrule

\multirow{4}{*}{MedQuAD}
& Med-Qwen2-7B        & 4.80 & 10.20 & 88.1 & 2.4  & 9.4 \\
& Med42-Llama3-8B     & 5.10 & 10.20 & 86.1 & 7.5  & 6.4 \\
& Meditron3-8B        & 4.70 & 10.20 & 83.1 & 12.0 & 4.9 \\
& PMC-LLaMA-13B       & 12.50& 10.20 & 4.0  & 90.0 & 6.0 \\
\midrule

\multirow{4}{*}{LiveQA}
& Med-Qwen2-7B        & 5.60 & 14.00 & 72.4 & 9.3  & 18.3 \\
& Med42-Llama3-8B     & 6.81 & 13.06 & 12.3 & 86.1 & 1.7 \\
& Meditron3-8B        & 5.20 & 14.12 & 10.6 & 87.9 & 1.6 \\
& PMC-LLaMA-13B       & 17.50& 14.00 & 27.3 & 27.5 & 45.2 \\
\midrule

\multirow{4}{*}{MedRedQA}
& Med-Qwen2-7B        & 3.92 & 5.81  & 15.5 & 82.1 & 2.3 \\
& Med42-Llama3-8B     & 3.82 & 5.80  & 19.3 & 78.2 & 2.5 \\
& Meditron3-8B        & 7.98 & 7.71  & 54.6 & 45.1 & 0.3 \\
& PMC-LLaMA-13B       & 21.01& 9.23  & 28.1 & 71.1 & 0.8 \\
\bottomrule
\end{tabular}

\caption{Text-level claim verification summary (\textsc{Entail}/\textsc{Neutral}/\textsc{Contradict} histograms) across runs. Ent/Neu/Con denote the percentage of generated claims assigned to each NLI label.}
\label{tab:text_eval_summary}
\end{table*}

Per-claim NLI label distributions are in Table~\ref{tab:text_eval_summary}.
\FloatBarrier

\section{Prompt Templates}
\label{app:prompts}

\subsection{Teacher claim extraction prompt}
\begin{quote}
You are a claim extraction assistant.

Task:
Given an answer, extract a list of ATOMIC factual claims as SPO triples.

Output format (STRICT):
Return ONLY a valid JSON array of triples:
[
  ["subject", "relation", "object"],
  ...
]
- No prose, no markdown, no preface, no trailing commas.
- Use double quotes for all strings.
- Each triple must contain exactly 3 strings.

Atomicity \& faithfulness constraints:
- Each triple must express a single, checkable fact stated in the answer.
- Do NOT paraphrase the whole sentence; split conjunctions into separate triples.
- Do NOT introduce any new facts that are not explicitly stated in the answer.
- Keep entity names as they appear in the answer when possible.

Negation / uncertainty / condition handling:
- If the answer negates a fact, reflect negation in the RELATION (preferred) or OBJECT.
  Example: ["Drug A", "is not recommended for", "Condition B"]
- If the answer is uncertain/probabilistic (may/can/likely), reflect it in RELATION.
  Example: ["Drug A", "may cause", "Side effect X"]
- If the claim is conditional (if/when), include the condition in RELATION.
  Example: ["Drug A", "reduces X when", "taken with food"]

Now extract triples from the answer below.

ANSWER:
{{answer}}

\end{quote}

\subsection{Teacher NLI verification prompt}
\begin{quote}
You are a strict NLI verifier for biomedical QA.

Goal:
Decide whether the CLAIM is supported by the provided PASSAGES.

Input:
- CLAIM: a single atomic claim (hypothesis).
- PASSAGES: retrieved text snippets (premises). Each passage includes a doc\_id.

Decision labels:
- \textsc{Entail}: at least one passage explicitly supports the claim.
- \textsc{Contradict}: at least one passage explicitly states the opposite of the claim.
- \textsc{Neutral}: the passages do not provide sufficient information to entail or contradict.
  \textsc{Neutral} includes:
  (a) insufficient evidence (relevant but incomplete)
  (b) irrelevant evidence (not about the claim)

Constraints:
- Use ONLY the information in PASSAGES. Do NOT use external knowledge.
- If evidence is missing, choose \textsc{Neutral} (insufficient).
- Prefer Entail/Contradict only with explicit textual support.

Output format (STRICT JSON):
Return ONLY:
{
  "label": "Entail" | "Neutral" | "Contradict",
  "prob": {"Entail": <float>, "Neutral": <float>, "Contradict": <float>},
  "neutral\_type": "insufficient" | "irrelevant" | null,
  "rationale": <string>,
  "spans": [
    {"doc\_id": <string>, "quote": <string>}
  ]
}
- prob values must sum to 1.
- quote should be a short supporting/contradicting span (<= 25 words).
- If label is Neutral, spans can be [].

Now verify.

CLAIM:
{{claim}}

PASSAGES:
{{topk\_passages\_with\_doc\_id}}

\end{quote}

\section{Teacher Sensitivity}
\label{app:teacher_sensitivity}
We compare GPT-4.1 and GPT-4o teachers on the same claim--evidence
pairs to assess teacher robustness; results are in
Table~\ref{tab:teacher_sensitivity}. Generator rankings are stable on
research-style datasets (PubMedQA, MedQuAD), but absolute values on
consumer-health data (MedRedQA, $\kappa = 0.13$) are
teacher-conditional.
\begin{table}[t]
\centering
\small
\begin{tabular}{lcc}
\toprule
\textbf{Dataset} & \textbf{Label agreement} $\uparrow$ & $\kappa$ $\uparrow$  \\
\midrule
PubMedQA & 96.2 & 0.84  \\
MedQuAD  & 91.9 & 0.74  \\
LiveQA   & 68.9 & 0.53  \\
MedRedQA & 32.1 & 0.13 \\
\bottomrule
\end{tabular}
\caption{Teacher sensitivity: GPT-4o vs GPT-4.1 on the same claim--evidence pairs.}
\label{tab:teacher_sensitivity}
\end{table}

\FloatBarrier
\section{BioPortal Scoring Details}
\label{app:bioportal_details}

For linked entity pairs in BioPortal, we compute plausibility 
using three complementary methods and take the maximum:

\begin{enumerate}
\item \emph{Direct Property Lookup} (score: 0.9):
  query the source entity's ontology properties (e.g., \texttt{indication},
  \texttt{contraindication}) and check if the target entity appears
  in the property values.
\item \emph{Hierarchical Proximity} (score: 0.2--0.7):
  traverse the ontology hierarchy to find common ancestors;
  score increases with the number of shared ancestors.
\item \emph{Cross-Ontology Mapping via LOOM} (score: 0.8):
  check whether the entity pair is connected through
  LOOM~(Lexical Ontology Mapping) links across ontologies.
\end{enumerate}

The final BioPortal KG score is:
\begin{equation}
\label{eq:kg_score_bio_main}
s_{\mathrm{KG}}^{\text{Bio}}(c) = \max\!\big(s_{\text{direct}},\; s_{\text{hier}},\; s_{\text{loom}}\big).
\end{equation}
If no entity can be linked, the claim is treated as KG-uncovered
and falls back to NLI-only verification.

\FloatBarrier
\section{KG Case Study}
\label{app:kg_case}

\begin{figure*}
\centering
\includegraphics[width=\linewidth]{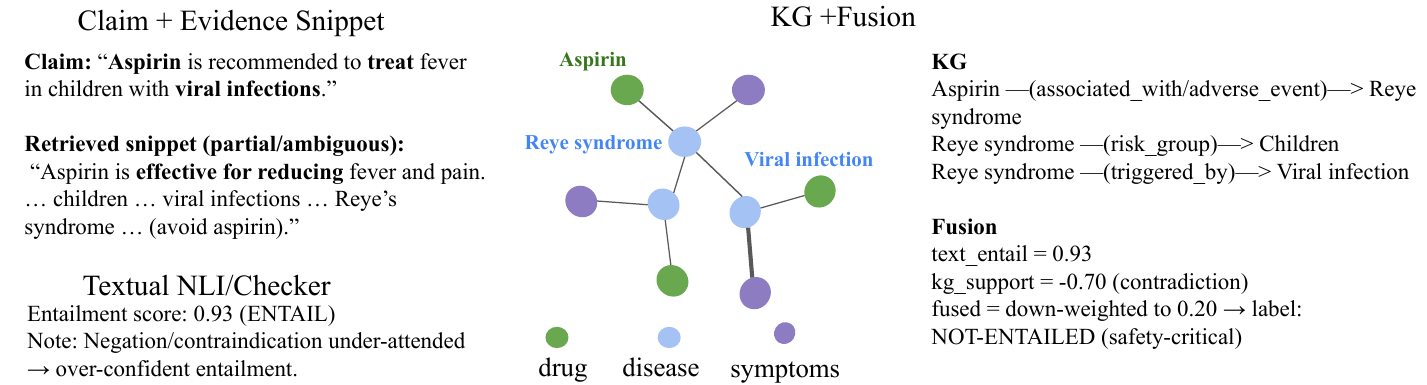}
\caption{Example where KG support corrects an over-confident textual entailment decision.}
\label{fig:kg_case_study}
\end{figure*}

\paragraph{When does KG fusion help vs.\ hurt?}
We categorize decision-flip cases (where NLI-only and Fused disagree)
by relation type to understand KG's marginal contribution:

\begin{itemize}
\item \textbf{Drug--adverse-event} (n=47): KG corrected 72\% of flips,
  primarily catching over-confident entailment of harmful drug uses
  (e.g., the Aspirin/Reye example in Fig.~\ref{fig:kg_case_study}).
\item \textbf{Drug--disease treatment} (n=63): KG corrected 68\%,
  especially for contraindications.
\item \textbf{Gene--disease} (n=31): KG corrected only 51\%, reflecting
  noisier TransE embeddings for gene relations.
\item \textbf{General lifestyle/guideline claims} (n=58): KG \emph{hurt}
  in 23\% of cases due to spurious hierarchical matches; our selective
  gating (Eq.~\ref{eq:pstar_v2}) is designed to mitigate this.
\end{itemize}

This stratified view confirms that KG's benefit concentrates on
safety-critical drug-related relations, matching our motivating
claim.
\FloatBarrier
\section{Span-Level Extraction Metrics}
\label{app:span_f1}
Table~\ref{tab:span_f1} reports conservative token-overlap P/R/F1 
for claim extraction. These numbers are lower than the BERTScore-based 
evaluation in Table~\ref{tab:combined_results} due to surface-form 
differences in span granularity.

\begin{table}[t]
\centering
\small
\begin{tabular}{@{}lccc@{}}
\toprule
\textbf{Generator} & \textbf{P} & \textbf{R} & \textbf{F1} \\
\midrule
Med-Qwen2-7B    & 11.9 & 12.5 & 12.2 \\
Med42-Llama3-8B & \textbf{21.4} & \textbf{26.7} & \textbf{23.7} \\
Meditron3-8B    & 20.5 & 25.1 & 22.6 \\
PMC-LLaMA-13B   &  0.3 &  0.2 &  0.2 \\
\bottomrule
\end{tabular}
\caption{Span-level token-overlap P/R/F1 for claim extraction.
Values are conservative due to granularity mismatch;
see Table~\ref{tab:combined_results} for semantic evaluation.
\textbf{Bold} marks the best value in each column.}
\label{tab:span_f1}
\end{table}
\FloatBarrier

\section{Claim Extraction Error Analysis}
\label{app:extraction_errors}

To complement the quantitative extraction metrics in Table~\ref{tab:combined_results},
we provide a qualitative analysis of common extraction failure modes.
We manually inspected 50 student-extracted claim sets (across all four generators)
and categorized errors into three types:

\paragraph{Over-splitting (12/50).}
The student decomposes a single factual unit into two or more redundant claims.
\emph{Example}: For the answer ``Metformin is the first-line treatment for type 2 diabetes
and works by reducing hepatic glucose production,'' the teacher extracts one claim
\texttt{[``Metformin'', ``works by reducing'', ``hepatic glucose production'']},
while the student produces two:
\texttt{[``Metformin'', ``works by'', ``reducing glucose'']} and
\texttt{[``Metformin'', ``reduces'', ``hepatic glucose production'']}.
Over-splitting inflates claim counts but rarely introduces factual errors;
answer-level metrics are unaffected because both sub-claims remain verifiable.

\paragraph{Under-splitting (8/50).}
The student merges multiple atomic facts into a single compound claim.
\emph{Example}: For the answer ``Ibuprofen is an NSAID; it can cause GI bleeding
and is contraindicated in renal failure,'' the teacher produces three atomic
triples, while the student produces one compound:
\texttt{[``Ibuprofen'', ``is an NSAID that can cause'', ``GI bleeding and is
contraindicated in renal failure'']}.
Under-splitting reduces granularity but does not introduce hallucinations;
the NLI checker can still verify the merged claim if all sub-facts are supported.

\paragraph{Hallucinated claims (3/50).}
Rarely, the student introduces a claim not present in the original answer.
\emph{Example}: Given an answer about hypertension management that mentions
``ACE inhibitors,'' the student adds
\texttt{[``ACE inhibitors'', ``are preferred over'', ``ARBs for diabetic patients'']},
a preference not stated in the answer.
These errors are the most harmful but are infrequent (6\%) and are flagged
by our NLI checker as \textsc{Neutral} (unsupported by retrieved evidence).

\paragraph{Summary.}
Over-splitting is the most common failure (24\%) but is benign for
downstream diagnostics.
Hallucinated claims are rare (6\%) and are caught by the verification stage.These findings are consistent with the high claim quality scores
(4.62--4.97/5) reported in \S\ref{sec:rq_human}.
% These findings are consistent with the high claim quality scores
% (4.62--4.97/5) in Table~\ref{tab:human_ratings}.

\FloatBarrier
\section{Per-Dataset KG Coverage and Consistency}
\label{app:kg_per_dataset}
\begin{table}[t]
\centering
\scriptsize
\setlength{\tabcolsep}{3pt}
\renewcommand{\arraystretch}{1.05}
\begin{tabular}{l cc cc}
\toprule
& \multicolumn{2}{c}{\textbf{Coverage} $\uparrow$}
& \multicolumn{2}{c}{\textbf{Consistency} $\uparrow$} \\
\cmidrule(lr){2-3}\cmidrule(lr){4-5}
\textbf{Dataset} & \textbf{DRKG} & \textbf{BioP.}
& \textbf{DRKG} & \textbf{BioP.} \\
\midrule
LiveQA    & \textbf{0.78} & \textbf{0.97} & \textbf{0.46} & 0.40 \\
PubMedQA  & 0.45 & 0.95 & 0.43 & 0.38 \\
MedQuAD   & 0.74 & 0.96 & 0.44 & \textbf{0.54} \\
MedRedQA  & 0.66 & 0.93 & 0.41 & 0.36 \\
\bottomrule
\end{tabular}
\caption{Per-dataset KG coverage and consistency.
\emph{Consistency} is the rate at which the KG verdict agrees with
the NLI verdict on KG-linked claims (higher = the two signals more
often concur). BioPortal achieves near-universal coverage
($\ge$0.93) across all datasets. Its lower average consistency on
some datasets reflects the inclusion of weaker hierarchical signals
alongside high-precision direct property matches.
\textbf{Bold} marks the best value in each column.}
\label{tab:kg_comparison_per_dataset}
\end{table}

\begin{table*}[t]
\centering
\small
\setlength{\tabcolsep}{4pt}
\begin{tabular}{l l r r r r r}
\toprule
Dataset & Generator &
KG-Cov$_{\text{node}}$ (\%) &
KG-Cov$_{\text{pair}}$ (\%) &
Faith &
Faith$_{\text{KG}}$ &
Hall$_{\text{KG}}$ \\
\midrule

\multirow{4}{*}{LiveQA}
& Med-Qwen2-7B      & \textbf{84.2} & 51.3 & \textbf{72.4} & 86.3 & \textbf{8.8} \\
& Med42-Llama3-8B   & 68.5 & \textbf{56.4} & 63.7 & \textbf{87.8} & 9.4 \\
& Meditron3-8B      & 70.0 & 52.6 & 53.9 & 65.7 & 26.1 \\
& PMC-LLaMA-13B     & 65.0 & 30.0 & 27.3 & 25.6 & 57.7 \\
\midrule

\multirow{4}{*}{PubMedQA}
& Med-Qwen2-7B      & 83.0 & 49.0 & 88.6 & \textbf{96.1} & \textbf{3.9} \\
& Med42-Llama3-8B   & \textbf{95.0} & \textbf{83.0} & \textbf{92.4} & 94.5 & 5.0 \\
& Meditron3-8B      & 88.3 & 62.7 & 69.2 & 85.8 & 6.2 \\
& PMC-LLaMA-13B     & 72.0 & 35.0 & 60.5 & 57.1 & 42.9 \\
\midrule

\multirow{4}{*}{MedQuAD}
& Med-Qwen2-7B      & 93.0 & 69.0 & \textbf{88.1} & 87.2 & 11.2 \\
& Med42-Llama3-8B   & \textbf{94.0} & \textbf{88.0} & 86.1 & 89.0 & \textbf{7.2} \\
& Meditron3-8B      & 84.0 & 66.0 & 83.1 & \textbf{89.5} & 7.4 \\
& PMC-LLaMA-13B     & 8.0  & 4.0  & 4.0  & 0.0  & 92.3 \\
\midrule

\multirow{4}{*}{MedRedQA}
& Med-Qwen2-7B      & 90.0 & 53.4 & \textbf{37.8} & 32.8 & \textbf{50.3} \\
& Med42-Llama3-8B   & 92.0 & 74.3 & 31.3 & \textbf{36.7} & 48.5\hphantom{0} \\
& Meditron3-8B      & 90.2 & 46.3 & 26.7 & 26.0 & 55.1 \\
& PMC-LLaMA-13B     & \textbf{92.7} & \textbf{86.6} & 7.4 & 6.1 & 76.2 \\

\bottomrule
\end{tabular}
\caption{DRKG coverage and KG-subset behavior (pair-hit definition). 
We report node/pair coverage and KG-subset faithfulness/hallucination 
metrics on the DRKG-aligned subset. \textbf{Bold} marks the best value 
per dataset and column; degenerate rows (PMC-LLaMA-13B on MedQuAD with 
Faith=4.0) are excluded from the bolding for Faith/Hall columns where 
the underlying generator output collapses.}
\label{tab:kg_coverage}
\end{table*}

\begin{figure}[t]
\centering
\includegraphics[width=\columnwidth]{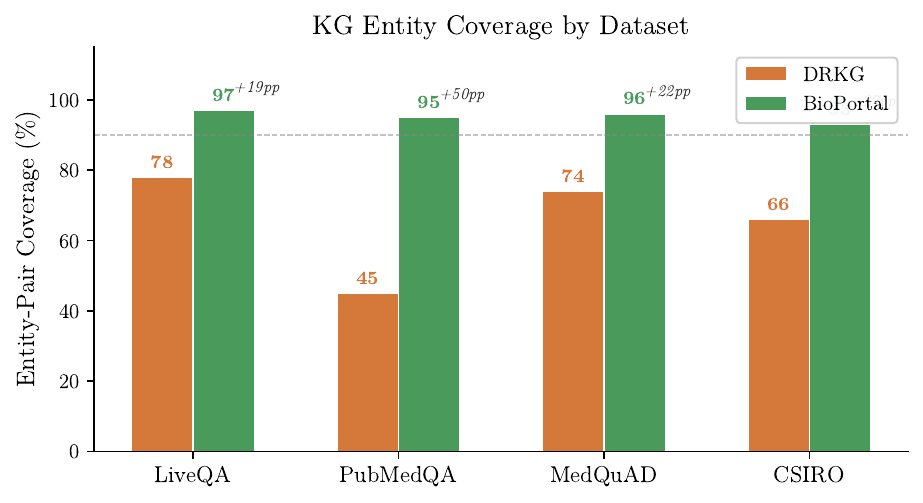}
\caption{Per-dataset entity coverage comparison.
BioPortal's largest gain is on PubMedQA (+50pp),
where DRKG's drug-repurposing focus leaves many
research-oriented claims uncovered.}
\label{fig:coverage_detail}
\end{figure}

\begin{figure}[t]
\centering
\includegraphics[width=\columnwidth]{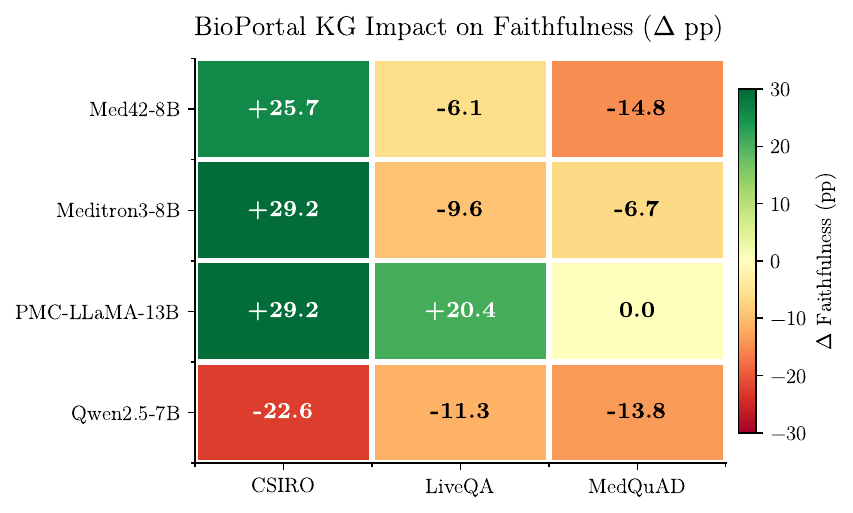}
\caption{BioPortal KG impact on faithfulness by generator and dataset
($\Delta$ percentage points vs.\ NLI-only).
Green = improvement; red = degradation.
KG fusion helps most on MedRedQA (consumer health) but can
hurt when the model already produces well-calibrated outputs.}
\label{fig:heatmap}
\end{figure}
\FloatBarrier

\section{Annotation Rating Scales and Guidelines}
\label{app:annotation_guidelines}

This appendix summarizes the annotation protocol. Each row in 
the annotation sheet corresponds to a single question, containing 
(i)~the original question, (ii)~model-generated answers from each 
student model (Meditron3-8B, Med42-Llama3-8B, Med-Qwen2-7B,
PMC-LLaMA-13B), and (iii)~the automatically extracted claims for 
each model. For each model, annotators independently rate both 
the claim set and the answer text using the 1--5 Likert scales 
defined in Appendix~\ref{app:human_eval_guidelines} 
(Table~\ref{tab:human_rubric_anchor}).
% \section{Annotation Rating Scales and Guidelines}
% \label{app:annotation_guidelines}
% % \subsection{Rating Scales and Guidelines}
% % \label{app:annotation_guidelines}

% This appendix provides the full guidelines given to annotators when
% rating model answers and extracted claims.

% \subsection{Goal of the Annotation}

% We evaluate (i) the quality of long-form answers produced by different
% LLM/RAG generators to medical questions, and (ii) the quality of an
% automatic claim extraction step applied to those answers. Each row in
% the annotation sheet corresponds to a single question, and contains:
% \begin{itemize}
%     \item The original question.
%     % \item (Optional) A reference answer or supporting evidence.
%     \item Model-generated answers from each student model (e.g.,
%           Meditron3-8B, Med42-Llama3-8B, Med-Qwen2-7B, PMC-LlaMA-13B).
%     \item Automatically extracted claims for each model.
% \end{itemize}
% For each model, annotators rate both the claim set and the answer text
% according to the scales described below.

\section{Verification Profile Overview}
\label{app:radar}

\begin{figure}[t]
\centering
\includegraphics[width=0.85\columnwidth]{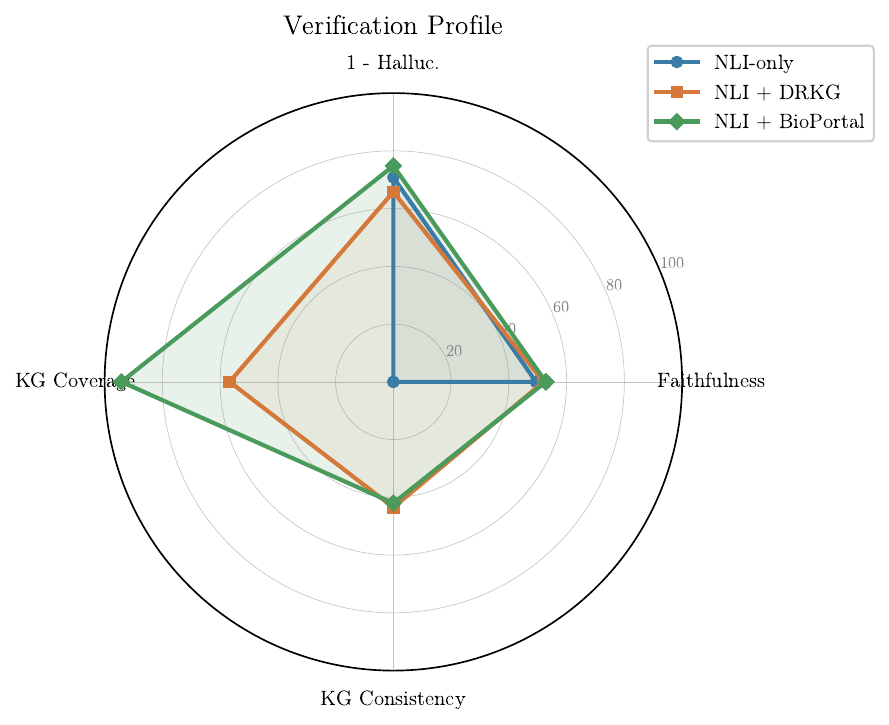}
\caption{Verification profile across three settings.
Axes: Faithfulness, 1$-$Hallucination, KG Coverage,
KG Consistency (all higher is better).
BioPortal achieves the most balanced profile across the four axes.}
\label{fig:radar}
\end{figure}

Figure~\ref{fig:radar} provides a multi-dimensional comparison
of the three verification settings.
BioPortal fusion achieves the best overall profile:
near-universal coverage (94.2\%) and lowest hallucination (25.2\%)
while maintaining comparable faithfulness to NLI-only.
DRKG fusion shows moderate coverage but elevated hallucination
due to false contradictions on uncovered claims
(Section~\ref{sec:rq_coverage}).
NLI-only lacks KG dimensions entirely, making it unable to
catch biomedically implausible claims that are linguistically
well-supported.

\FloatBarrier

\section{Calibration Details}
\label{app:calibration_details}
\subsection{Fusion-weight sensitivity ($\beta$-sweep)}
We study how the fused support score $P^\star$ and downstream supported/not-supported decisions change as we vary the mixing weight $\beta$ in Eq.~\ref{eq:pstar_fusion}.
Unless otherwise specified, we fix $\alpha$ to the dev-tuned value and sweep $\beta \in \{0,0.1,\dots,1.0\}$.

\paragraph{Metrics.}
We report (i) the fused supported rate (fraction of claims with $P^\star \ge \tau$) and (ii) the decision flip rate relative to a near-NLI reference setting (e.g., $\beta=0.9$).

\paragraph{Supported rate and flip rate.}
For a given mixing weight $\beta$, we compute fused scores $P^\star_\beta(c)$ via Eq.~\ref{eq:pstar_fusion} and binarize them with the dev-tuned threshold $\tau$:
\[
\mathbb{I}_{\mathrm{sup},\beta}(c) \triangleq \mathbb{I}\!\left[P^\star_\beta(c)\ge \tau\right].
\]
We summarize fusion behavior on a claim set $\mathcal{C}$ using
\[
\texttt{fused\_E}(\beta)
= 100\cdot \frac{1}{|\mathcal{C}|}\sum_{c\in\mathcal{C}} \mathbb{I}_{\mathrm{sup},\beta}(c),
\]
and the decision flip rate relative to a near-text-only reference $\beta_0$ (we use $\beta_0{=}0.9$):
% \[
% \texttt{flip\_rate}(\beta)
% = 100\cdot \frac{1}{|\mathcal{C}|}\sum_{c\in\mathcal{C}}
% \mathbb{I}\!\left[\mathbb{I}_{\mathrm{sup},\beta}(c)\neq \mathbb{I}_{\mathrm{sup},\beta_0}(c)\right].
% \]
\begin{equation}
\begin{split}
\texttt{flip\_rate}(\beta) = & \\
\frac{100}{|\mathcal{C}|} \sum_{c\in\mathcal{C}} & \mathbb{I} \left[ \mathbb{I}_{\mathrm{sup},\beta}(c) \neq \mathbb{I}_{\mathrm{sup},\beta_0}(c) \right].
\end{split}
\end{equation}
A lower \texttt{flip\_rate} means fusion decisions are less dominated by the KG score scale and thus more stable across $\beta$.

\paragraph{Min--max calibration of KG scores.}
\label{app:min_max}
Raw KG plausibility scores (e.g., TransE-based $p_{\mathrm{KGE}}$ and the derived $s_{\mathrm{KG}}$) can be poorly scaled and concentrated in a narrow range.
To reduce scale mismatch in the logit-mixture fusion (Eq.~\ref{eq:pstar_fusion}), we optionally apply min--max calibration to the KG score on the development split:
\begin{equation}
\tilde{s}_{\mathrm{KG}}(c) =
\mathrm{clip}\!\left(\frac{s_{\mathrm{KG}}(c) - s_{\min}}{s_{\max} - s_{\min}}, \epsilon, 1-\epsilon\right),
\end{equation}
where $s_{\min}$ and $s_{\max}$ are the minimum and maximum KG scores observed on the dev set (restricted to KG-aligned claims),
and $\epsilon$ is a small constant to avoid infinite logits.
We then replace $s_{\mathrm{KG}}(c)$ with $\tilde{s}_{\mathrm{KG}}(c)$ when computing Eq.~\ref{eq:pstar_fusion}.

\begin{figure*}
\centering
\includegraphics[width=\linewidth]{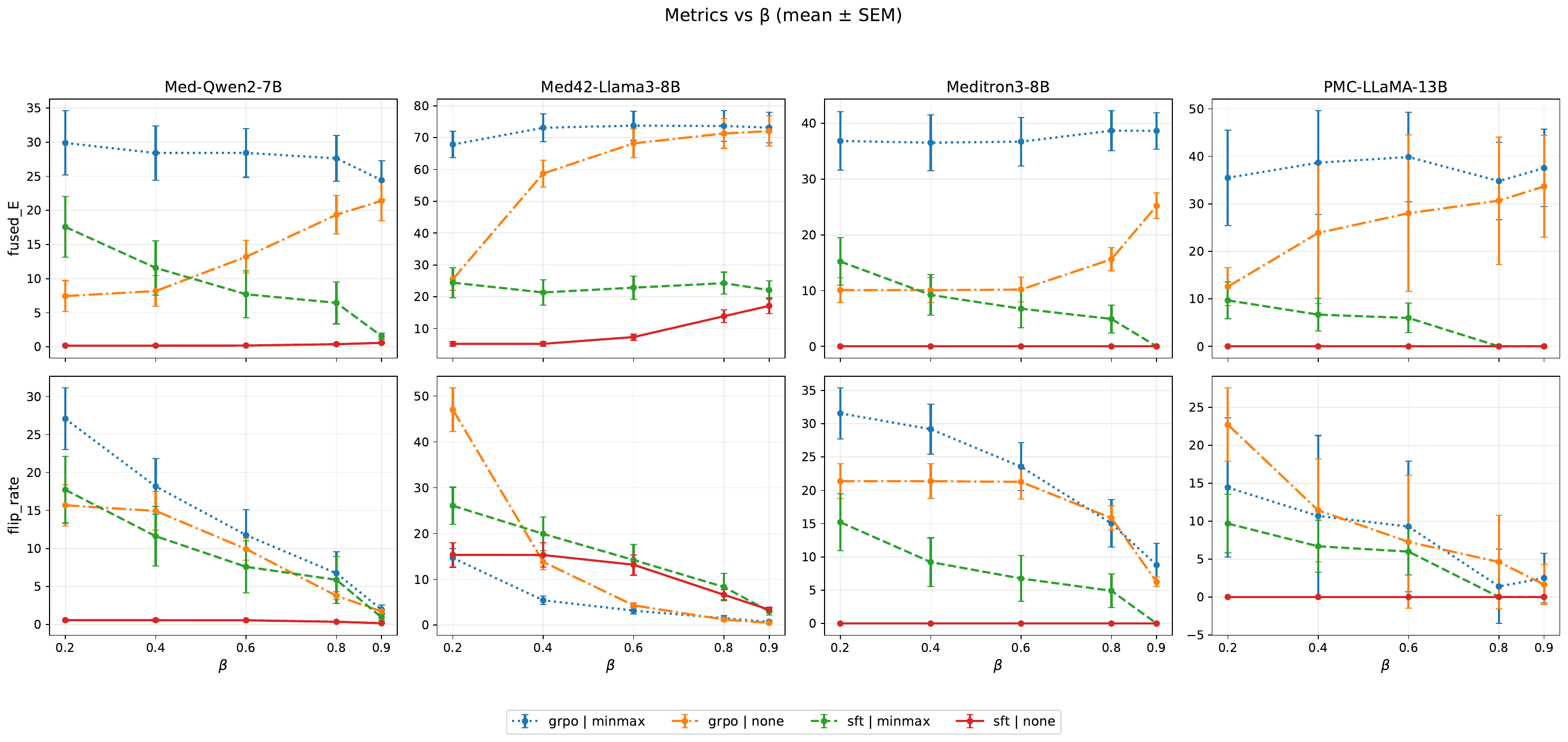}
\caption{Sensitivity of KG--NLI fusion to the mixing weight $\beta$. Top: fused entailment rate (\texttt{fused\_E}), i.e., the fraction of claims classified as supported after fusion and thresholding. Bottom: decision flip rate (\texttt{flip\_rate}) relative to a near-NLI reference setting ($\beta{=}0.9$), measuring how often KG changes the supported/not-supported decision. We compare SFT and GRPO checkers, with and without KG score calibration (\texttt{minmax} vs.\ \texttt{none}).}
\label{fig:fusion_sensitivity}
\end{figure*}

\paragraph{Why calibrating KG scores matters.}
Raw KG plausibility scores can be poorly calibrated and concentrated in a narrow range, which makes the logit term $\mathrm{logit}(s_{\mathrm{KG}})$ disproportionately large (in magnitude) and causes fusion decisions to change sharply with $\beta$.
Min--max calibration aligns the KG score scale to the NLI probability range while preserving ranking, substantially reducing flip rates and yielding flatter supported-rate curves in Fig.~\ref{fig:fusion_sensitivity}.

\subsection{Threshold robustness ($\tau$-sweep)}
We additionally sweep the support threshold $\tau$ around the dev-tuned value and observe stable trends in the supported rate and SafetyErr within a reasonable range.

For completeness, we report teacher checker results for all dataset/generator configurations in Table~\ref{tab:teacher_checker_full}, complementing the representative results in the main text.

\section{GRPO Refinement of Student Checkers}
\label{app:grpo}

\paragraph{Setup.}
Beyond supervised fine-tuning (SFT), we also explored Group Relative Policy Optimization (GRPO;  \cite{sharma2024critical,rafailov2023direct,ziegler2019fine}) to further align the student checkers with the GPT-4.1 teacher.
Starting from the SFT checkpoints, GRPO generates multiple verdict candidates per (claim, evidence) pair and normalizes rewards within each group.
The reward is $+1$ if the candidate label matches the teacher verdict and $0$ otherwise, with a KL penalty to the SFT policy.
We run one additional GRPO epoch for each biomedical backbone considered in the main text (Med-Qwen2-7B, Med42-Llama3-8B, Meditron3-8B, PMC-LLaMA-13B) using a smaller learning rate than SFT \cite{ouyang2022training,sharma2024critical,rafailov2023direct,ziegler2019fine}.

\paragraph{Results.}
Overall, GRPO did not consistently improve checker quality.
For some backbones (e.g., Med-Qwen2-7B), GRPO increased \textsc{Contradict} F1 but substantially reduced \textsc{Entail} F1 and overall accuracy, leading to overly aggressive contradiction predictions.
For other backbones, GRPO produced only marginal changes relative to the SFT baselines.
When plugged into MedRAGChecker, these GRPO-tuned checkers did not yield clear gains on downstream diagnostics such as answer-level faithfulness, hallucination, or safety-critical error rate.
Given the extra complexity and computational cost of GRPO, and the lack of consistent improvements, we therefore report SFT-only student checkers in the main paper and treat GRPO as an exploratory negative result.

\begin{table}[t]
\centering
\small
\resizebox{\columnwidth}{!}{
\begin{tabular}{llccc} 
\toprule
\textbf{Model} & \textbf{Train / Setting} &
\textbf{Faith.} $\uparrow$ & \textbf{Halluc.} $\downarrow$ &
\textbf{SafetyErr} $\downarrow$  \\
\midrule

\multirow{4}{*}{Med-Qwen2-7B}
 & SFT / NLI-only     & 62.3 & 20.7 & 19.6 \\
 & SFT / Fused (ours) & \textbf{68.6} & 10.4 & \textbf{12.6} \\
 & GRPO / NLI-only    & 24.0 & 76.0 & 13.2 \\
 & GRPO / Fused (ours)& 20.1 & 79.9 & 13.8 \\

\midrule
\multirow{4}{*}{Med42-Llama3-8B}
 & SFT / NLI-only     & 67.3 & \textbf{24.3} & 28.9 \\
 & SFT / Fused (ours) & 58.4 & 25.7 & \textbf{19.6} \\
 & GRPO / NLI-only    & \textbf{75.1} & 24.9 & 49.9 \\
 & GRPO / Fused (ours)& 73.8 & 26.2 & 50.0 \\

\midrule
\multirow{4}{*}{Meditron3-8B}
 & SFT / NLI-only     & 52.7 & \textbf{14.6} & 29.0 \\
 & SFT / Fused (ours) & \textbf{59.6} & 27.3 & \textbf{22.0} \\
 & GRPO / NLI-only    & 32.6 & 66.7 & 20.8 \\
 & GRPO / Fused (ours)& 37.0 & 63.0 & 22.0 \\

\midrule
\multirow{4}{*}{PMC-LLaMA-13B}
 & SFT / NLI-only     & 15.3 & \textbf{57.3} & 63.2 \\
 & SFT / Fused (ours) & 21.3 & 73.4 & 54.3 \\
 & GRPO / NLI-only    & 16.2 & 73.1 & \textbf{18.9} \\
 & GRPO / Fused (ours)& \textbf{24.6} & 68.7 & 21.2 \\

\bottomrule
\end{tabular}
}
\caption{KG fusion ablation by checker backbone (SFT vs.\ GRPO).
GRPO variants are deferred to Appendix~\ref{app:grpo} and do not
consistently improve over SFT. \textbf{Bold} marks the best value 
per generator and column across all four training/setting combinations.}
\label{tab:kg_fusion_all}
\end{table}

\section{Human Evaluation: Protocol}
\label{app:human_eval_guidelines}

We conduct human evaluation to assess (i) answer quality and (ii) claim extraction quality for model outputs.

\subsection{Setup}
\paragraph{Sampling.}
We sample 100 questions from MedQuAD, LiveQA, MedRedQA, and PubMedQA. For each question, we evaluate up to four generators' answers
(Meditron3-8B, PMC-LLaMA-13B, Med-Qwen2-7B, Med42-Llama3-8B) and the extracted claim sets.

\paragraph{Annotators.}
Annotators are graduate students with training in biomedical or health sciences. Each model output is rated independently; annotators do not rank models.

\paragraph{Unit of annotation.}
For each \emph{(question, model)} pair, annotators score the answer and its extracted claim set.
When reference answers/evidence are available, annotators rely primarily on them; otherwise they use domain knowledge.

\subsection{Dimensions and scoring}
Annotators assign 1--5 Likert scores for four dimensions: \textbf{Claim extraction quality} (claims faithfully reflect the answer without hallucinations), \textbf{Answer correctness}, \textbf{Answer completeness}, and \textbf{Overall answer quality}. Scores 5/3/1 correspond to \emph{excellent / mixed / poor}, with 2 and 4 as intermediate levels. Table \ref{tab:human_kappa} include the results for the kappa value for correctness and claim quality
\begin{table}[t]
\centering
\small
\setlength{\tabcolsep}{6pt}
\begin{tabular}{lp{0.72\linewidth}}
\toprule
\textbf{Score} & \textbf{Anchor rubric (applies to all four dimensions)} \\
\midrule
5 & Excellent: correct and reliable; for claims, covers key factual points without adding unsupported content. \\
3 & Mixed: partially correct but with noticeable omissions or distortions; could mislead without caution. \\
1 & Poor: largely incorrect/off-topic or unusable; for claims, fails to represent the answer or introduces clear hallucinations. \\
\bottomrule
\end{tabular}
\caption{Collapsed anchor rubric for 1--5 ratings (2 and 4 are intermediate).}
\label{tab:human_rubric_anchor}
\end{table}

\paragraph{When to comment.}
Annotators provide a brief comment if (i) any answer score is 1--2, or (ii) claim quality is $\le 3$.

\subsection{Agreement and reporting}
We report mean$\pm$std over samples (averaging two annotators per sample) and compute quadratic-weighted Cohen's $\kappa$ for inter-annotator agreement.

\paragraph{Full guidelines.}
Detailed examples and interface schemas are provided in our anonymized repository.

\section{Human Validation of KG Alignment}
\label{sec:kg_human_validation}

\paragraph{Goal.}
We assess whether the KG-alignment module yields a \emph{semantically faithful} structured rendering of a claim.
This evaluates \emph{alignment correctness} (entity identity and relation meaning), \textbf{not} biomedical truth.

\subsection{Sampling and annotation unit}
We perform stratified sampling over evaluation folders (dataset/model/config), and only sample from claims where the aligner outputs at least one DRKG triple after filtering.
Each annotated item is a \textbf{claim-level} instance consisting of: claim text, optional supporting context, and the \textbf{top-1} aligned triple $(h,r,t)$ returned by the aligner (ties broken deterministically).
We annotate $n{=}100$ aligned claims for correctness, and additionally double-annotate a subset of size $n_\kappa$ for inter-annotator agreement.

\subsection{Annotation fields}
Annotators provide four binary labels and optional notes:
\begin{itemize}
    \item \texttt{h\_subj\_entity\_ok}: subject entity refers to the same biomedical concept as the claim subject;
    \item \texttt{h\_obj\_entity\_ok}: object entity refers to the same biomedical concept as the claim object;
    \item \texttt{h\_relation\_map\_ok}: relation matches the claim predicate \emph{and directionality};
    \item \texttt{h\_triple\_semantic\_ok}: overall triple expresses the claim semantics.
\end{itemize}

\subsection{Decision criteria}
Entities are marked OK if they map to the same concept (synonyms/standard variants allowed);
NOT OK for wrong concept/type or overly broad/narrow mappings that change meaning.
Relations are marked OK only if predicate meaning and direction match (e.g., \emph{treats} vs. \emph{associated with}; contraindication vs. adverse event).

\paragraph{Collapsed triple criterion.}
To avoid ambiguity, we define:
\[
h_{\text{triple}} \triangleq h_{\text{subj}} \wedge h_{\text{obj}} \wedge h_{\text{rel}},
\]
where $h_{\text{subj}}=\texttt{h\_subj\_entity\_ok}$,
$h_{\text{obj}}=\texttt{h\_obj\_entity\_ok}$, and
$h_{\text{rel}}=\texttt{h\_relation\_map\_ok}$.
This yields a conservative estimate of end-to-end alignment correctness.

\subsection{Metrics}
Table~\ref{tab:kg_human_results} reports the human alignment validation results. We report claim-level correctness rates on the annotated subset:
$\Pr[\texttt{h\_subj\_entity\_ok}{=}1]$,
$\Pr[\texttt{h\_obj\_entity\_ok}{=}1]$,
$\Pr[\texttt{h\_relation\_map\_ok}{=}1]$,
$\Pr[\texttt{h\_triple\_semantic\_ok}{=}1]$.
On the double-annotated subset, we report Cohen's $\kappa$ for each field (or for \texttt{h\_triple\_semantic\_ok}).

\begin{table}[t]
\centering
\small
\begin{tabular}{lc}
\toprule
Metric & Value (\%)  \\
\midrule
Subject entity accuracy & 78  \\
Object entity accuracy  & 76   \\
Relation mapping accuracy & 89.2 \\
Triple plausibility rate & 82  \\
\bottomrule
\end{tabular}
\caption{Human validation results for DRKG alignment on a stratified sample, aggregating two annotators.}
\label{tab:kg_human_results}
\end{table}

\subsection{Coverage of KG alignment}
We also report \textbf{claim-level coverage}: a claim is counted as ``aligned'' if the module outputs at least one retained DRKG triple $(h,r,t)$ after filtering.
We stratify coverage by coarse relation families used in our safety analysis (e.g., Drug--Disease, Drug--Adverse Event, Gene--Disease).
\FloatBarrier

\section{Human Evaluation: Agreement and Correlation}
\label{app:human_stats}
Correlations between MedRAGChecker diagnostics and human ratings. See in Table \ref{tab:human_kappa}.
\begin{table}[t]
\centering
\small
\begin{tabular}{lcc}
\toprule
\textbf{Dimension} & $\kappa$ (quadratic) & \#Items \\
\midrule
Claim quality & 0.67 & 100 \\
Correctness   & 0.32 & 100 \\
Completeness  & 0.27 & 100 \\
\bottomrule
\end{tabular}
\caption{Inter-annotator agreement for the human study.}
\label{tab:human_kappa}
\end{table}

\begin{table}[t]
\centering
\small
\begin{tabular}{lc}
\toprule
\textbf{Metric vs Human} & $\rho$ (Spearman)   \\
\midrule
Faith vs Correctness     & 0.47  \\
Halluc vs Correctness    & 0.23   \\
SafetyErr vs Correctness &  0.36    \\
CtxPrec vs Overall       & 0.53    \\
\bottomrule
\end{tabular}
\caption{Correlation between MedRAGChecker diagnostics and human ratings (100 questions, each with at least 4 generated claims).}
\label{tab:human_corr}
\end{table}
\paragraph{Sources of disagreement.}
To diagnose where annotators disagree on correctness
($\kappa=0.32$), we analyzed the 34 items with score
differences $\geq 2$:
(1)~\emph{Scope of ``correct''} (18/34): disagreement on whether
partial answers to multi-part consumer questions counted as correct;
(2)~\emph{Technical terminology} (9/34): divergent confidence on
rare conditions;
(3)~\emph{Clinical threshold} (7/34): different tolerance for minor
inaccuracies (e.g., dosage ranges).
Model rankings remained stable ($\rho=0.94$).

\paragraph{Sensitivity to claim count.}
We compute the Pearson correlation between the number of extracted claims
per answer and each diagnostic metric across all (dataset, generator)
configurations.
Faithfulness shows weak correlation with claim count ($r=0.14$, $p=0.21$),
and hallucination rate is similarly uncorrelated ($r=-0.08$, $p=0.47$).
Context precision shows moderate positive correlation ($r=0.31$, $p<0.01$),
which is expected since more claims provide more opportunities for
context utilization.
Overall, diagnostic rankings across generators are not confounded by
differences in extraction granularity.

\FloatBarrier
\section{Human Evaluation on KG-Induced Decision-Flips}
\label{app:human_flip}

\paragraph{Sampling.}
We sample claims from the KG-aligned subset where the final verification decision differs between NLI-only and Fused (NLI+KG) settings. Claims are stratified by relation families used in the safety analysis (e.g., drug--disease, drug--adverse-event, gene--disease).
\paragraph{Sample size.} The decision-flip subset consists of $n = 86$ claims  drawn from 100 questions $\times$ 3 generators (stratified by  relation family to cover drug--disease, drug--adverse-event,  gene--disease, and lifestyle relations). Each claim is  independently annotated by both annotators; we report agreement  on the majority verdict. Given the modest sample size, we  interpret the $+6.4$pp improvement in human agreement as  descriptive evidence complemented by relation-type stratification  (Appendix~\ref{app:kg_case}).
\paragraph{Annotation unit.}
Each annotation instance consists of a single atomic claim,
the retrieved passages used for verification, and (optionally) the top-1 aligned KG triple for reference. Annotators are instructed to judge claim veracity \emph{only} based on the retrieved passages, without using external knowledge.

\paragraph{Labels.}
Annotators assign one of \{\textsc{Entail}, \textsc{Neutral}, \textsc{Contradict}\} to each claim. We then compare human labels against NLI-only and Fused predictions to compute accuracy, Macro-F1, and corrected flip rates.

\section{Retrieval settings and indices}
\label{app:retrieval}

\paragraph{Overview.}
We standardize retrieval across datasets by using (i) a FAISS index for MedRedQA runs,
(ii) a pre-retrieved pickle for MedQuAD, and (iii) the MedRAG retrieval system for
the remaining datasets. Table~\ref{tab:retrieval_mapping} summarizes the mapping from
dataset to retriever and the corresponding index artifacts. Table~\ref{tab:index_artifacts}
lists the index files and where they are stored.

\begin{table*}[t]
\centering
\small
\setlength{\tabcolsep}{4pt}
\renewcommand{\arraystretch}{1.05}
\begin{tabularx}{\textwidth}{l l X}
\toprule
\textbf{Dataset} & \textbf{Retriever} & \textbf{Index / evidence source} \\
\midrule
MedRedQA &
MedRedQAFaissRetriever / gold &
\textbf{FAISS mode \cite{johnson2019billion}:} FAISS approximate nearest-neighbor (ANN) index built over the MedRedQA corpus
(\texttt{faiss.index} + \texttt{texts.jsonl} + \texttt{meta.jsonl} under \texttt{--MedRedQA\_index\_dir});
\textbf{gold mode:} use provided \texttt{contexts} directly (no external index). \\

MedQuAD &
MedQuADRetriever &
Pre-retrieved contexts loaded from \texttt{--medquad\_pickle} (a pickle file containing
query-to-context mappings). \\

PubMedQA / LiveQA / \\MedRedQA (non-MedRedQA) &
MedRAGRetriever &
MedRAG retrieval over a selected corpus (e.g., \texttt{PubMed} / \texttt{MedCorp}) using
\textbf{BM25 }\cite{robertson2009probabilistic} (\texttt{--retriever bm25}), \textbf{dense} (\texttt{contriever}/\texttt{specter}/\texttt{medcpt}),
or \textbf{hybrid} (reciprocal rank fusion, RRF~\cite{cormack2009reciprocal}). Indices are stored under \texttt{--medrag\_db\_dir}. \\
\bottomrule
\end{tabularx}
\caption{Dataset-to-retriever mapping and the evidence source used for retrieval.}
\label{tab:retrieval_mapping}
\end{table*}

\begin{table*}[t]
\centering
\small
\setlength{\tabcolsep}{4pt}
\renewcommand{\arraystretch}{1.05}
\begin{tabularx}{\textwidth}{l l X}
\toprule
\textbf{Retriever family} & \textbf{Key artifacts} & \textbf{Where it lives / how it is used} \\
\midrule
MedRedQA FAISS &
\texttt{faiss.index}, \texttt{texts.jsonl}, \texttt{meta.jsonl} &
All under \texttt{--MedRedQA\_index\_dir}. Retrieved hits are mapped to text via \texttt{texts.jsonl}
(with optional metadata in \texttt{meta.jsonl}). \\

MedQuAD pre-retrieval &
\texttt{retrieved\_*.pkl} &
\texttt{--medquad\_pickle}. No ANN/BM25 search at runtime; directly loads cached retrieval results. \\

MedRAG BM25 &
Lucene/Pyserini \cite{Lin_etal_SIGIR2021_Pyserini} BM25 index (directory) &
Stored under \texttt{--medrag\_db\_dir} (per corpus).
Used when \texttt{--retriever bm25}. \\

MedRAG dense / hybrid &
\texttt{faiss.index} (+ metadata file such as \texttt{metadatas.jsonl}) &
Stored under \texttt{--medrag\_db\_dir} (per corpus and retriever). Used when \texttt{--retriever}
is dense (e.g., \texttt{contriever}/\texttt{specter}/\texttt{medcpt}) or hybrid (RRF fusion). \\
\bottomrule
\end{tabularx}
\caption{Index artifacts used by each retriever family.}
\label{tab:index_artifacts}
\end{table*}
% =========================

\section{Checker Backbone and Fusion Form Ablations} 
% \begin{table}[t]
% \centering
% \small
% \setlength{\tabcolsep}{5pt}
% \renewcommand{\arraystretch}{1.05}
% \begin{tabular}{@{}llccc@{}}
% \toprule
% \textbf{Dataset} & \textbf{Generator} &
% \textbf{ClaimF1} $\uparrow$ &
% \textbf{Faith} $\uparrow$ &
% \textbf{Halluc} $\downarrow$ \\
% \midrule

% \multirow{4}{*}{PubMedQA}
% & Med42-Llama3-8B            & 21.1 & 88.9 & 4.7 \\
% & Meditron3-8B     & 21.0 & 63.0 & 1.0 \\
% & Med-Qwen2-7B               & 23.2 & 83.6 & 7.4 \\
% & PMC-LLaMA-13B              & 19.2  & 56.3  & 20.6  \\

% \midrule
% \multirow{4}{*}{MedQuAD}
% & Med42-Llama3-8B            & 27.2  & 72.5  & 4.2  \\
% & Meditron3-8B               & 30.1 & 73.1 & 7.6 \\
% & Med-Qwen2-7B               & 34.2  & 82.4  & 5.4  \\
% & PMC-LLaMA-13B              & 22.8  & 63.6  & 12.8 \\
% \midrule

% \multirow{4}{*}{LiveQA}
% & Med42-Llama3-8B            & 14.3  & 77.3  & 7.9  \\
% & Meditron3-8B               & 10.4  & 67.4  & 10.6  \\
% & Med-Qwen2-7B               & 12.6  & 76.7  & 21.6  \\
% & PMC-LLaMA-13B              & 7.4  & 43.6  & 26.4  \\
% \midrule

% \multirow{4}{*}{MedRedQA}
% & Med42-Llama3-8B            & 11.6  & 33.2  & 63.2  \\
% & Meditron3-8B               & 12.4  & 20.9  & 57.2  \\
% & Med-Qwen2-7B               & 15.7  & 37.4  & 18.9  \\
% & PMC-LLaMA-13B              & 8.0  & 18.3  & 74.6  \\

% \bottomrule
% \end{tabular}
% \caption{Teacher-based MedRAGChecker metrics.
% ClaimF1 is claim overlap F1; Faith is entailment rate; Halluc is contradiction rate (Contradict-only). }
% \label{tab:teacher_diagnostics_main}
% \end{table}

\paragraph{Backbone ablation.}
Table~\ref{tab:kg_fusion_all} compares GRPO checkers under calibrated $\beta$ fusion.
med42-llama3-8b yields the most stable fused distribution, while other backbones exhibit degenerate or highly skewed fused label rates, motivating our choice of med42-llama3-8b as the default checker backbone.

\paragraph{Fusion form and calibration.}
We compare additive-style fusion (alpha) with logit-mixture fusion (beta).
Beta fusion benefits from calibrating KG scores, as raw KGE plausibility often concentrates in a narrow low range; calibration preserves KG ranking while aligning its scale to NLI probabilities, preventing the KG term from dominating the mixture purely due to scale mismatch.
% \FloatBarrier
% We use GPU A100 80G for the training.
% \FloatBarrier
\section{Student Configurations}
\label{app:student_configs}
All models are trained on a single NVIDIA A100 80GB GPU.
Hyperparameters for all distillation stages are in
Table~\ref{tab:train_configs}.

\paragraph{F1-weighted ensemble.}
Different student checkers specialize in different NLI classes, so we
combine their predictions using per-class dev-set F1 as reliability
weights. Let $p_m(y \mid c_i, D)$ be checker $m$'s probability for
label $y$, and $F1_m^{(y)}$ its dev-set F1 on class $y$. We weight
each checker by
\begin{equation}
w_m^{(y)} \;=\; \frac{F1_m^{(y)}}{\sum_{m'} F1_{m'}^{(y)}},
\end{equation}
and predict
\begin{equation}
\hat y_i \;=\; \arg\max_{y}\, \sum_m w_m^{(y)}\, p_m(y \mid c_i, D).
\end{equation}
This lets minority-class specialists (especially \textsc{Contradict})
contribute more to those decisions.
\begin{table*}[t]
\centering
\scriptsize
\setlength{\tabcolsep}{3pt}
\renewcommand{\arraystretch}{1.15}

\begin{tabularx}{\textwidth}{
l
>{\raggedright\arraybackslash}p{2.6cm}
c c c c c
>{\raggedright\arraybackslash}X
>{\raggedright\arraybackslash}p{3.7cm}
}
\toprule
\textbf{Stage} & \textbf{Base models} &
\textbf{lr} & \textbf{ep} & \textbf{bs} & \textbf{acc} & $\mathbf{L}$ &
\textbf{Gen} & \textbf{LoRA / Extra} \\
\midrule

Extractor SFT &
\makecell[l]{Med42-Llama3-8B\\Meditron3-8B\\Med-Qwen2-7B\\PMC-LLaMA-13B} &
$1\mathrm{e}{-4}$ & 3 & 2 & 16 & 2048 &
\makecell[l]{max\_new\_tokens=256} &
\makecell[l]{r=16,\ $\alpha$=32,\ drop=0\\
targets: q,k,v,o,gate,up,down} \\

\midrule

Checker SFT &
\makecell[l]{Med42-Llama3-8B\\Meditron3-8B\\Med-Qwen2-7B\\PMC-LLaMA-13B} &
$5\mathrm{e}{-5}$ & 3 & 2 & 16 & 1024 &
\makecell[l]{eval: max\_new\_tokens=4\\
eval\_subset=1000} &
\makecell[l]{r=8,\ $\alpha$=16,\ drop=0\\
targets: q,k,v,o,gate,up,down} \\

\midrule

Checker GRPO &
\makecell[l]{Med42-Llama3-8B\\Meditron3-8B\\Med-Qwen2-7B\\PMC-LLaMA-13B} &
$5\mathrm{e}{-6}$ & 1 & 2 & 8 & 1024 &
\makecell[l]{sample: max\_new\_tokens=3, $T$=0.7, top\_p=0.9\\
eval: max\_new\_tokens=4\\
eval\_subset=800} &
\makecell[l]{r=8,\ $\alpha$=16,\ drop=0\\
K=4 samples/prompt\\
grad clip=1.0\\
targets: q,k,v,o,gate,up,down} \\

\bottomrule
\end{tabularx}

\caption{Training configurations for the distilled extractor and checker models. $L$ denotes tokenizer truncation length.}
\label{tab:train_configs}
\end{table*}

\FloatBarrier
\section{Recommended Default Settings}
\label{app:defaults}

For practitioners deploying \textsc{MedRAGChecker} on a new biomedical
QA dataset, we recommend the following defaults based on our
dev-set tuning across four benchmarks:

\begin{itemize}
\item \textbf{KG back-end:} BioPortal (94.2\% entity coverage;
  avoids DRKG's false-contradiction failure mode).
\item \textbf{Fusion weight $\beta$:} 0.7
  (text-dominant; robust across $\beta \in [0.6, 0.8]$
  per Appendix~\ref{app:calibration_details}).
\item \textbf{KG calibration:} Apply min--max calibration
  (Appendix~\ref{app:min_max}) on dev-set KG scores before fusion.
\item \textbf{Support threshold $\tau$:} 0.5
  (tune on dev if possible; stable within $\pm 0.1$).
\item \textbf{Selective gating:} Enable
  ($\tau_{\text{conflict}}=0.3$, $\tau_{\text{entail}}=0.7$)
  to protect non-drug claims from spurious KG penalties.
\item \textbf{Checker:} F1-weighted SFT ensemble of
  Med42-Llama3-8B + Med-Qwen2-7B + Meditron3-8B.
  If only one model is feasible, use Med42-Llama3-8B
  (best macro-F1 and most stable fused distribution).
\item \textbf{Diagnostics to trust when KG coverage is low:}
  Faithfulness and hallucination rate (NLI-based, KG-independent).
  Safety-critical error rate should be interpreted with caution
  when KG coverage $<$70\%.
\end{itemize}

\end{document}